\documentclass[11pt]{article}

\usepackage[final]{acl}

\usepackage{times}
\usepackage{latexsym}
\usepackage{xcolor,colortbl}
\usepackage{colortbl,array}
\usepackage{colortbl}
\usepackage{colortbl,array,xcolor}
\usepackage{booktabs}
\usepackage{xcolor}
\usepackage{tikz}
\usetikzlibrary{tikzmark}
\usepackage{amsmath}
\usepackage{amssymb}
\usepackage{arydshln}
\usepackage{multirow}
\usepackage{algorithm}
\usepackage{algpseudocode}   
\usepackage{xcolor}
\usepackage{subfigure}
\usepackage{placeins}
\usepackage{caption}
\usepackage{hyperref}

\usepackage{tcolorbox}

\newcommand*\myfontsize{%
  \@setfontsize\myfontsize{7}{8}%
}

\makeatother
\newcommand{\mytextbox}[2]{%
  \colorbox{#1!15}{\strut #2}%
}

\definecolor{myred}{HTML}{E74C3C}
\definecolor{myblue}{HTML}{3498DB}
\definecolor{mygreen}{HTML}{27AE60}
\definecolor{myorange}{HTML}{ff8800}
\definecolor{mypurple}{HTML}{9B59B6}
\definecolor{mypink}{HTML}{f7acb9}

\definecolor{headercolor}{RGB}{100,100,100}
\definecolor{bluecolor}{RGB}{0,100,200}
\definecolor{orangecolor}{RGB}{200,100,50}
\definecolor{redcolor}{RGB}{200,0,0}

\newcommand{\red}[1]{\mytextbox{myred}{\textbf{\textcolor{myred}{#1}}}}
\newcommand{\blue}[1]{\mytextbox{myblue}{\textbf{\textcolor{myblue}{#1}}}}
\newcommand{\green}[1]{\mytextbox{mygreen}{\textbf{\textcolor{mygreen}{#1}}}}

\newcommand{\purple}[1]{\mytextbox{mypurple}{\textbf{\textcolor{mypurple}{#1}}}}

\usepackage[T1]{fontenc}

\usepackage[utf8]{inputenc}

\usepackage{microtype}

\usepackage{inconsolata}

\usepackage{graphicx}

\title{AT\textsuperscript{2}PO: Agentic Turn-based Policy Optimization via Tree Search}

\author{
Zefang Zong{$^{1}\thanks{~~Equal contribution.}$}, Dingwei Chen{$^{1,2}\footnotemark[1]$}, Yang Li{$^{1}$}, Qi Yi{$^{1}$}, Bo Zhou{$^{1}$}, \\ \textbf{Chengming Li{$^{3}$}, Bo Qian{$^{1}$}, Peng Chen{$^{1}$}, Jie Jiang{$^{1}\thanks{~~Corresponding author.}$}} \\
 $^1$Tencent Inc $^2$ Sun Yat-Sen University $^3$Shenzhen MSU-BIT University
\\ \texttt{\{willzong,thomasyngli\}@tencent.com} \\
\texttt{cuso4cdw@gmail.com, licm@smbu.edu.cn}
  }

\begin{document}

\maketitle
\begin{abstract}
LLM agents have emerged as powerful systems for tackling multi-turn tasks by interleaving internal reasoning and external tool interactions. Agentic Reinforcement Learning has recently drawn significant research attention as a critical post-training paradigm to further refine these capabilities. In this paper, we present AT\textsuperscript{2}PO (\textbf{A}gentic \textbf{T}urn-based \textbf{P}olicy \textbf{O}ptimization via \textbf{T}ree Search), a unified framework for multi-turn agentic RL that addresses three core challenges: limited exploration diversity, sparse credit assignment, and misaligned policy optimization. AT\textsuperscript{2}PO introduces a turn-level tree structure that jointly enables \textit{Entropy-Guided Tree Expansion} for strategic exploration and \textit{Turn-wise Credit Assignment} for fine-grained reward propagation from sparse outcomes. Complementing this, we propose \textit{Agentic Turn-based Policy Optimization}, a turn-level learning objective that aligns policy updates with the natural decision granularity of agentic interactions. ATPO is orthogonal to tree search and can be readily integrated into any multi-turn RL pipeline.  Experiments across seven benchmarks demonstrate consistent improvements over the state-of-the-art baseline by up to 1.84 percentage points in average, with ablation studies validating the effectiveness of each component. Our code is available at \href{https://github.com/zzfoutofspace/ATPO}{\texttt{https://github.com/zzfoutofspace/ATPO}}.
\end{abstract}

\section{Introduction}

Recent advances in large language models (LLMs) have catalyzed the development of autonomous agents capable of executing complex, multi-turn tasks through structured reasoning and tool use \cite{yao2023react, schick2023toolformer, shen2024smartcal}. Reinforcement learning (RL) offers a natural framework for refining these agents by optimizing behavior toward task-level success. Inspired by recent progress in RL with Verifiable Rewards (RLVR) which leverages programmatic feedback to guide training \cite{guo2025deepseek, wang2025reinforcementlearningreasoninglarge, wen2025reinforcementlearningverifiablerewards, lambert2025tulu3pushingfrontiers}, Agentic RL extends this paradigm to interactive, multi-turn scenarios. In this context, agents act through turns of interleaved thinking and tool use, with tool responses forming the subsequent observations, creating a structured, turn-based interaction that requires specialized algorithmic designs. This emerging paradigm of agentic RL has drawn significant attention in the research community \cite{jin2025search, wu2025webdancer, dong2025agenticaepo, feng2025group, yan2025memory}.
\begin{figure}[t]
	\centering
\includegraphics[width=0.47\textwidth]{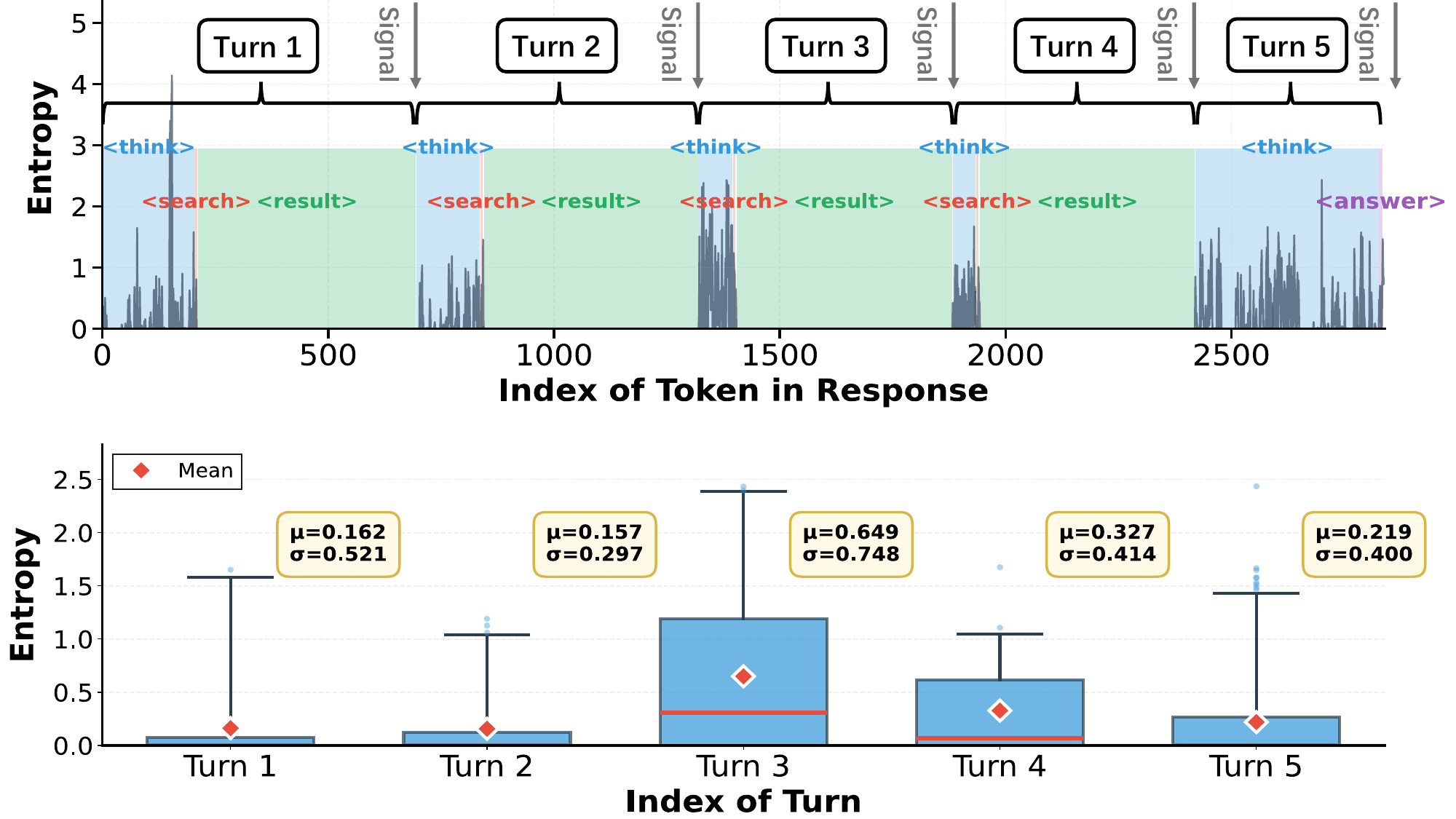}
    \vspace{-2mm}
	\caption{\textit{Above}: Token-level entropy distribution in a search agentic rollout sample. A complete rollout sequence is sampled from multiple turns. \textit{Below}: Turn-wise entropy box plot analysis. Substantial discrepancies exist among turn-level token groups. }
    \label{fig:box plot}
    \vspace{-4mm}
\end{figure}

Despite its promise, Agentic RL still faces three fundamental challenges that hinder effective policy learning. First, agents struggle to generate high-quality, diverse trajectories under limited rollout budgets. Existing frameworks typically rely on chain-based generation or tree-based rollouts that expand nodes using random or heuristic criteria. These methods fail to strategically prioritize the exploration of high-uncertainty or high-potential turns, thereby limiting trajectory diversity and quality. Second, learning signals in multi-turn trajectories are inherently sparse. Rewards are typically available only at the end of a complete trajectory, making it difficult to attribute credit to specific intermediate action steps. \textbf{Most critically, there remains a fundamental misalignment between the turn-based structure of agentic tasks and the flat optimization objectives used in current policy learning methods.} Current policy optimization algorithms inherited from RLVR treat the agent’s output as a flat token sequence by simply masking out the tool response tokens, and apply either token-level~\cite{shao2024deepseekmathpushinglimitsmathematical} or sequence-level~\cite{zheng2025groupsequencepolicyoptimization} importance sampling ratios. However, both schemes overlook a key structural property: the trajectory is generated through interleaved turns of agent reasoning and tool invocation, with new tool-generated tokens dynamically inserted between segments, as illustrated in Figure~\ref{fig:box plot}. Consequently, neither token-level nor sequence-level importance weighting properly accounts for the turn-wise sampling process. Moreover, even when step-wise learning signals are available, the unit of optimization mismatched with the natural unit of turn-level supervision. This structural mismatch prevents existing methods from capturing the stepwise logic of multi-turn decision-making, leading to unstable gradients and inefficient learning.

To overcome these challenges, we propose AT\textsuperscript{2}PO (\textbf{A}gentic \textbf{T}urn-based \textbf{P}olicy \textbf{O}ptimization via \textbf{T}ree Search), a unified framework that integrates three synergistic components tailored to multi-turn agentic RL. 1) In the rollout phase, we introduce \textit{Entropy-Guided Tree Expansion}, which adaptively grows the search tree from the most uncertain turns to maximize exploration efficiency. 2) In the rewarding phase, we develop a \textit{Turn-wise Credit Assignment} mechanism that propagates sparse outcome rewards backward through the tree to compute fine-grained, per-turn value and advantage estimates. It enables precise credit allocation without auxiliary supervision. 3) In the training phase, we design \textit{Agentic Turn-based Policy Optimization (ATPO)}, a novel policy learning algorithm that operates its importance sampling and clipping at the turn level, aligning the optimization objective with the structured turn-based decision process and significantly improving training stability and performance. Together, these components enable AT\textsuperscript{2}PO to generate more diverse and high-quality rollouts, leverage sparse rewards more effectively, and optimize policies in a manner faithful to the multi-turn agentic paradigm. In conclusion, our contributions are summarized as follows:
\begin{itemize}
\item We propose a unified tree-based approach that jointly leverages Entropy-Guided Tree Expansion for strategic exploration and Turn-wise Credit Assignment for fine-grained reward propagation. The turn-level tree maximizes exploration diversity and supervision quality from sparse outcome rewards.

\item We propose Agentic Turn-based Policy Optimization (ATPO), a turn-level policy update mechanism that is orthogonal to the tree-search framework and can be readily integrated into any multi-turn agentic RL pipeline. By operating at the natural granularity of multi-turn interactions, ATPO improves optimization stability and alignment compared to token-level or sequence-level baselines.

\item Through extensive experiments on search agent benchmarks, we demonstrate that AT\textsuperscript{2}PO consistently outperforms prior strong baselines, yielding gains of up to 1.84 percentage points in average over the state-of-the-art baseline across seven benchmarks. Ablation studies further validate the effectiveness of each component above.

\end{itemize}

\begin{figure*}[t]
	\centering
    \includegraphics[width=1\linewidth]{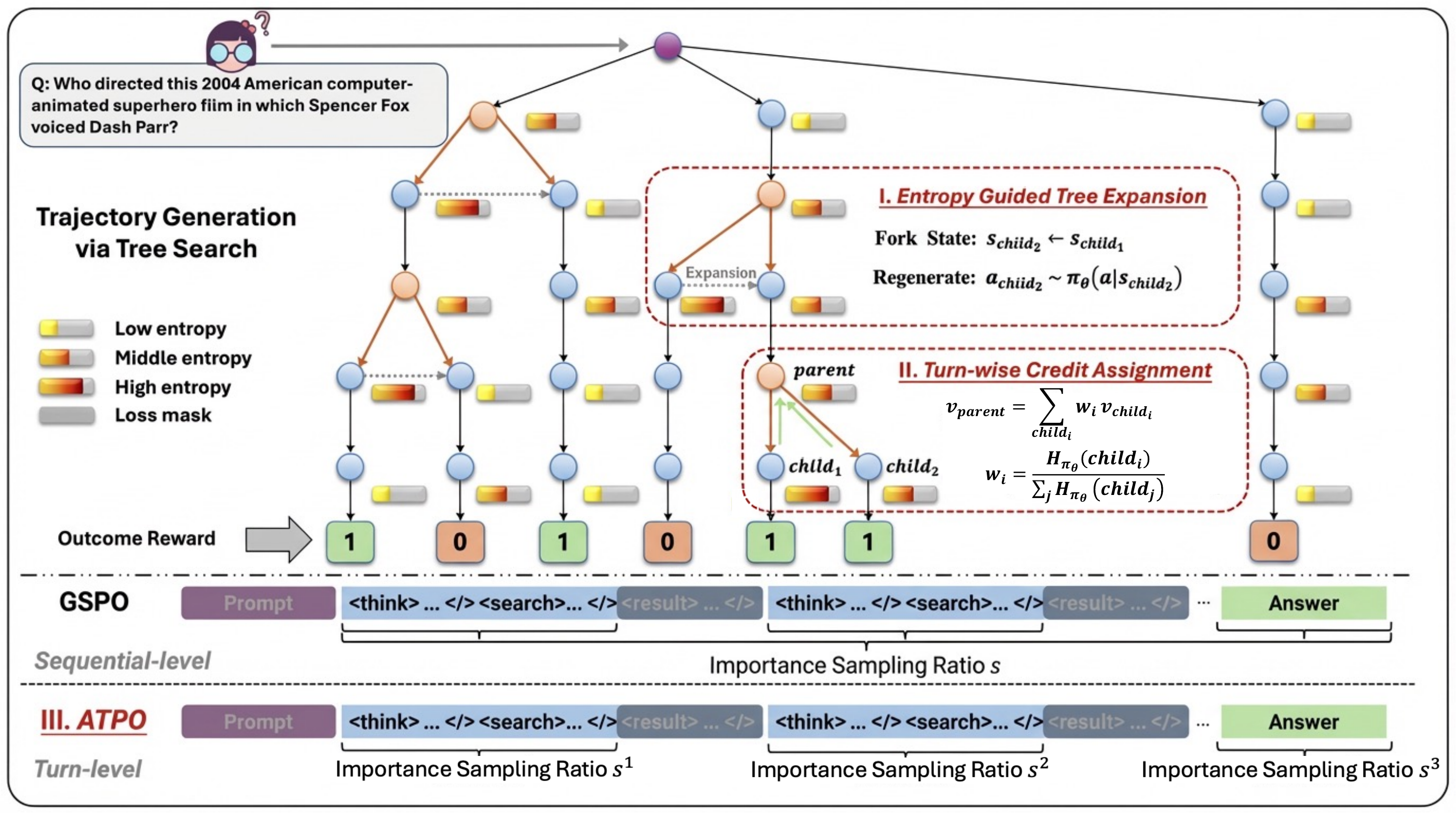}
    
	\caption{Overview of the \textbf{AT\textsuperscript{2}PO} framework, including entropy-guided tree-structured rollout, turn-wise credit assignment for fine-grained supervision and turn-based policy optimization during reinforcement training.}
	\label{f2}
    \vspace{-4mm}
\end{figure*}

\section{Related Work}

\subsection{Reinforcement Learning for LLMs}
RL has become a key approach for aligning LLMs with complex objectives and improving reasoning capabilities~\cite{yu2025dapo,zheng2025groupsequencepolicyoptimization,jin2025search,wang2025ragen}. Traditional RLHF pipelines rely on Proximal Policy Optimization (PPO)~\cite{schulman2017proximal} with a learned reward model and value critic~\cite{christiano2017deep,stiennon2020learning}. More recently, methods like RLVR leverage verifiable, programmatic rewards, demonstrating strong performance in settings with deterministic feedback~\cite{guo2025deepseek}. To simplify the architecture and reduce variance, a line of work replaces the critic with group-based baselines. GRPO~\cite{guo2025deepseek} computes relative advantages within response groups, achieving strong results on long-form reasoning. DAPO~\cite{yu2025dapo} improves scalability through decoupled clipping and dynamic sampling, while GSPO~\cite{zheng2025groupsequencepolicyoptimization} shifts clipping from the token to the sequence level for enhanced stability. Concurrent efforts further stabilize training via adaptive clipping strategies in the policy objective~\cite{chen2025minimax,wang2025aspo,su2025gppo,gao2025softadaptivepolicyoptimization}.

However, existing methods treat rollouts as flat sequences and do not explicitly account for the multi-turn nature of agentic interactions. In contrast, ATPO formulates the learning objective at the turn level, aligning policy updates with turn-wise feedback to improve training stability and 
quality.

\subsection{Agentic Reinforcement Learning}
Recent studies on Agentic RL focus primarily on trajectory data \cite{wu2025webdancer, li2025websailor}, algorithmic optimization \cite{dong2025agenticarpo, dong2025agenticaepo, feng2025group}, memory management \cite{yan2025memory, yu2025memagent}, as well as multimodality \cite{geng2025webwatcher, wu2025mmsearch} and efficiency \cite{wang2025otc}. Algorithmically, several works focus on structured exploration. Search-R1~\cite{jin2025search} integrates search tools into RL training, establishing a paradigm for knowledge-augmented agents. ARPO~\cite{dong2025agenticarpo} branches rollouts at high-entropy tokens to encourage exploration, while AEPO~\cite{dong2025agenticaepo} limits over-branching via entropy-aware clipping. GiGPO~\cite{feng2025group} introduces multi-granularity advantage estimation for finer credit assignment. Most closely related, Tree-GRPO~\cite{ji2025treesearchllmagent} organizes rollouts into trees and computes group-relative advantages both within and across branches, extending GRPO to agentic settings.

Even though methods like Tree-GRPO proposed to adopt a tree structure for more efficient rollout and trajectory collection, they do not explicitly prioritize expansion. In contrast, our entropy-guided tree expansion actively selects high-entropy nodes for branching, maximizing exploration efficiency. Coupled with ATPO our approach ensures coherent, diverse, and high-quality agentic behavior.

\section{Preliminaries}


An agentic framework interacts with external tools over multiple steps, which we model as a Markov Decision Process (MDP). At each time step $t$, the agent observes a state $s_t$ and selects an action $a_t$, resulting in a transition to the next state $s_{t+1}$.

Following the widely adopted ReAct paradigm~\cite{yao2023react}, each action 
$a_t$ comprises two components, including an internal chain-of-thought reasoning step and an executable operation. This operation is either a request to invoke an external tool or a final answer returned to terminate the interaction. Given a question in the input prompt, the agent iteratively samples actions $a_t\sim\pi_\theta(\cdot|s_t)$ according to its current policy $\pi_\theta$, continuing until it emits the final answer at step $T-1$. The resulting interaction trajectory is thus represented as $\tau=\{(s_t,a_t)\}_{t=0}^{T-1}$.

We assign a trajectory-level reward $R(\tau)$ based on the outcome, and aim to maximize the expected return as follows:

\vspace{-2mm}
\begin{equation}
J(\pi_\theta)=\mathbb{E}_{\tau\sim \pi_\theta}[R(\tau)],
\end{equation}

Optimization is typically performed via policy gradient methods. In contrast to actor-critic approaches such as PPO~\cite{schulman2017proximal}, our method follows group-based optimization~\cite{guo2025deepseek}, which leverages group-wise candidate sampling to implicitly estimate advantages without requiring an additional critic network.

\section{Methods}
In this section, we detail the methodology of AT\textsuperscript{2}PO, which introduces specific enhancements across the rollout, rewarding, and training phases. First, during the rollout phase, AT\textsuperscript{2}PO adopts a tree-based structure incorporating Entropy Guided Tree Expansion to enhance exploration diversity within a fixed inference budget. Second, in the rewarding phase, we leverage this tree topology to implement turn-wise credit assignment, enabling fine-grained supervision derived solely from outcome rewards. Finally, in the training phase, we design the turn-based policy optimization tailored for multi-turn agentic RL to enhance training stability and quality.

\subsection{Entropy Guided Tree Expansion}

During the rollout phase, tree-search strategies have demonstrated strong effectiveness in online agentic RL~\cite{ji2025treesearchllmagent}. In contrast to conventional RLVR settings where tree nodes correspond to specially selected individual tokens~\cite{hou2025treerl}, multi-turn agentic tasks naturally lend themselves to a turn-level formulation. Each observation–action pair constitutes a single tree node. This clear contextual segmentation facilitates diverse action sampling (including reasoning steps and tool-use requests) from well-structured intermediate states. 

Rather than relying on random or heuristic expansion strategies as in prior work (e.g., Tree-GRPO~\cite{ji2025treesearchllmagent}), we propose Entropy-Guided Tree Expansion, which iteratively expands the search tree from the most uncertain turns to promote diverse exploration. The full tree construction procedure is summarized in Algorithm~\ref{alg:tree-search-entropy}.
\begin{algorithm}[h]
\caption{Entropy Guided Tree Expansion}
\label{alg:tree-search-entropy}
\begin{algorithmic}[1]
\Require Prompts $\mathcal{P}$, initial branches $M$, beam size $K$, expansion iterations $L$, policy $\pi_\theta$, branching penalty coeff $\alpha$
\Ensure Tree structures $\mathcal{T} = \{T_1, T_2, \ldots, T_{|\mathcal{P}|}\}$

\Statex \textcolor{gray}{\textit{// Phase 1: Initialize Trees}}
\For{each prompt $x_i \in \mathcal{P}$}
    \State Initialize root node $n_i$ with state $s_{i,0} = x_i$
    \For{$m = 1$ to $M$}
        \State $\tau_m \gets \texttt{RolloutFrom}(s_{i,0};\pi_\theta)$
        \State Add all nodes $(s_{i,j}, a_{i,j})$ from $\tau_m$ to $T_i$
    \EndFor
\EndFor

\Statex \textcolor{gray}{\textit{// Phase 2: Node Sampling \& Tree Expansion}}
\For{$l = 1$ to $L$}
    \For{each tree $T_i$}
        \State $\mathcal{C} \gets \texttt{get\_non\_leaf\_nodes}(T_i)$ 
        \State $s(n) = H_{\pi_\theta}(n) - \alpha|\text{children}(n.parent)|$
        \State $\mathcal{S} \gets \texttt{TopK}(\mathcal{C}, K, s)$ 
        \For{each selected node $(s_{i,j}, a_{i,j}) \in \mathcal{S}$}
            \State $\tau' \gets \texttt{RolloutFrom}(s_{i,j};\pi_\theta)$
            \State Add all nodes $(s_{i,j}', a_{i,j}')$ from $\tau'$ to tree $T_i$ as new branch
        \EndFor
    \EndFor
\EndFor

\State \Return $\mathcal{T}$

\end{algorithmic}
\end{algorithm}

\textbf{Phase 1: Initialization.} For each prompt $x_i \in \mathcal{P}$ in the training batch, we first generate $M$ independent chain-based trajectories using the current policy $\pi_\theta$. Each prompt $x_i$ is associated with its own tree $T_i$ initialized by adding these $M$ trajectories as branches. All branches are rooted at a shared root node $n_i$ storing the initial state $s_{i,0} = x_i$. All non-root nodes contain both the intermediate state and the corresponding action, i.e., the incremental rollout segment generated at that turn.

\textbf{Phase 2.1: Sampling.} We compute the policy entropy at each node to quantify decision uncertainty. Higher entropy indicates greater ambiguity in action selection, highlighting nodes that would benefit most from further exploration. Specifically, we use Monte-Carlo estimation to estimate the normalized entropy $H_{\pi_\theta}(n)$ of each node $n$:
\vspace{-1mm}
\begin{equation}
\begin{split}
    H_{\pi_\theta}(n) =&\mathbb{E}_{y\sim\pi_\theta(\cdot|x)}[-\text{log}\pi_\theta(y^{k}|x, y^{<k})] \\
    \approx& \frac{1}{|y^k|}\sum_{y_t\in y^k}-\text{log}\pi_\theta(y_t|x, y^{<k})
\end{split}
\end{equation}
where $y^k$ denotes the rollout segment generated in the $k$-th turn in a trajectory and stored at node $n$. We use the superscript $y^k$ to indicate the subsequence corresponding to a single turn, and the subscript $y_t$ to refer to the individual tokens within that subsequence.

Based on these entropy estimates, we select the top $K$ highest-entropy nodes in each tree and fork their states $s_{i,j}$ to initiate new branches. To encourage balanced exploration across the tree and prevent over-expansion from a single node, we introduce a branching penalty coefficient $\alpha$ that downweights repeatedly selected nodes.

\textbf{Phase 2.2: Expansion.}
From each forked state $s_{i,j}$, we regenerate the remainder of the trajectory using the full context from the root to the fork point. The resulting continuation is appended as a new branch to the original tree.

We repeat Phase 2.1 and 2.2 for $L$ iterations yielding a total of $M+LK$ leaf nodes per prompt. As the expected depth of the sampled nodes for expansion is half of the maximum length, given that the average token budget of an individual rollout as $B$, the total token budget for one search tree is $(M+LK/2)B$. Thus the tree-based approach generates a larger number of diverse rollouts under the same computational budget, enhancing both exploration efficiency and training signal quality.

\subsection{Turn-wise Credit Assignment}
During the rewarding stage, most existing RLVR algorithms solely relies on the final outcome reward~\cite{shao2024deepseekmathpushinglimitsmathematical} as ground-truth verification is typically only feasible upon completion of the full response. This sparse credit assignment obscures the contribution of individual actions to final outcomes, yielding weak or delayed learning signals that impede effective credit propagation in multi-turn agentic tasks.

Leveraging the search tree constructed during rollout, AT\textsuperscript{2}PO enables fine-grained, turn-wise credit assignment. Specifically, we estimate the value $V_n$ of each node $n$ via Monte Carlo bootstrapping over its descendants,
\vspace{-1mm}
\begin{equation}
\begin{split}
V_n = & 
\begin{cases}
\hat{r}_n, & \text{if } n \text{ is leaf} \\
 \sum_{c \in C(n)} w_cV_c, & \text{otherwise,}
\end{cases} \\
w_c = & \frac{H_{\pi_\theta}(c)}{\sum_{c' \in C(n)}H_{\pi_\theta}(c')},
\end{split}
\end{equation}
where $C(n)$ denotes the set of child nodes of $n$, and $\hat{r}_n=\frac{r_n-mean\{r_n\}}{std\{r_n\}}$ is the normalized final outcome reward assigned via standard GRPO normalization only at leaves. This recursive aggregation effectively propagates the outcome-based rewards backward through the tree to each intermediate node.

Based on the node values, we compute turn-wise advantages $A_n=f(V_n)$ to serve as credit signals for policy updates, and assign them to all tokens within. Several $f(\cdot)$ strategies are viable, including treating $V_n$ directly as advantages, or other formulations proposed~\cite{hou2025treerl}. We evaluate the effectiveness of different credit assignment schemes in Section~\ref{sec:ablation-credit}.

\subsection{Agentic Turn-based Policy Optimization}
During training, we observe that existing agentic RL frameworks typically adopt optimization algorithms originally designed for conventional RLVR settings, such as GRPO in Tree-GRPO and GPPO~\cite{su2025klear} in AEPO. These methods treat the entire interaction as a monolithic sequence and apply standard policy gradients, without accounting for the unique structure of multi-turn agentic interactions.

Motivated by the success of recent sequence-level policy optimization methods like GSPO, which adjusts the learning objective into the sequence level to align with the final outcome reward, we propose Agentic Turn-Based Policy Optimization (ATPO) to better address the multi-turn nature of agentic tasks. Building on the fine-grained, turn-wise credit assignments derived from previous designs, we reformulate the policy optimization objective at the turn level for better alignment. Specifically, we introduce turn-wise importance sampling ratios and clipping mechanisms that operate per decision turn, enabling more precise and stable policy updates. This turn-aware formulation ensures that each reasoning or action step is (a) assigned importance ratios in a turn-consecutive manner and (b) optimized using its own turn-level learning signal.
\vspace{-2mm}
\begin{equation}
\begin{split}
& \mathcal{J}_{ATPO}(\theta) = \mathbb{E}_{x \sim \mathcal{D}, \{y_i\}_{i=1}^G \sim \pi_{\theta_{old}}(\cdot | x)} \Bigg[ \\
& \quad \frac{1}{G} \sum_{i=1}^G \frac{1}{|y_i|} \sum_{t=1}^{|y_i|} M_{i,t}\min \Bigg( s_{i,t}^{turn}(\theta) \hat{A}_{i,t}, \\
& \qquad \operatorname{clip} \left( s_{i,t}^{turn}(\theta), 1 - \varepsilon_l, 1 + \varepsilon_r \right) \hat{A}_{i,t} \Bigg) \Bigg],
\end{split}
\label{eq:gspo-turn}
\end{equation}
\begin{equation}
\begin{split}
 & s_{i,t}^{turn}(\theta) = \frac{\pi_\theta(y_{i,t} | x, y_{i,<t})}{\operatorname{sg} \left[ \pi_\theta(y_{i,t} | x, y_{i,<t}) \right]} \\
 & \quad \cdot \operatorname{sg} \left[ \Bigg( \frac{\pi_{\theta}(y_i^{k(t)}|x, y_i^{<k(t)})}{\pi_{\theta_{old}}(y_i^{k(t)}|x, y_i^{<k(t)})} \Bigg)^{\frac{1}{|y_i^{k(t)}|}} \right],
\end{split}
\label{eq:turn-importance-ratio}
\end{equation}
where the individual token $y_{i,t}$ belongs to the turn $y_i^{k(t)}$, and $sg(\cdot)$ indicates the stop gradient operation. $M_{i,t}$ is a masking scheme that includes only tokens generated by the inference engine, excluding those originating from tool responses.

The effectiveness of ATPO stems from its balanced design that bridges the gap between token-level and sequence-level policy optimization. First, in contrast to token-level methods such as GRPO, which suffer from high-variance of individual importance sampling ratios, ATPO operates at the turn level, significantly reducing clipped tokens and improving training stability. Second, unlike GSPO which facilitates its optimization objective and reward both on the coarse sequence-level, ATPO ensures that the optimization objective remains locally aligned with each consecutive turn-wise sampling step and with the fine-grained supervision provided by turn-based credit assignment.

Note that ATPO is orthogonal to the tree search framework, and serves as a plug-in design for any multi-turn agentic rl tasks. To better assess the effectiveness of turn-level optimization in ATPO, we also introduce a diagnostic metric called turn entropy $\mathcal{H}_{turn}$, which quantifies the degree of variation in policy updates across different turns within a trajectory. The details are presented in Appendix~\ref{sec:formulas}.

\section{Evaluation}

\begin{figure*}[t]
\captionsetup{type=table}  \small
\setlength\tabcolsep{8.5pt}
\renewcommand\arraystretch{1.2}
\begin{center}
\begin{tabular}{lccccccccc}

\toprule
\multirow{2}{*}{\textbf{Method}}&\multicolumn{5}{c}{\textbf{Multi-Hop QA}}&\multicolumn{4}{c}{\textbf{Single-Hop QA}}  \\
\cmidrule(r){2-6}  \cmidrule(r){7-10} 
& \textbf{Hotpot} & \textbf{2wiki} & \textbf{Musiq} & \textbf{Bamb} & \textbf{Avg.} & \textbf{NQ} & \textbf{TriviaQA} & \textbf{PopQA} & \textbf{Avg.}\\
\hline 
\rowcolor{pink!20}\multicolumn{10}{c}{\textbf{Backbone Model: Qwen3-4B}} \\
ReAct  &30.42	&32.92	&12.83	&44.80   &  30.01    &26.75	&53.53	&35.34 & 41.31\\
$+$ GRPO  &44.76 & 51.40 & 21.60 & 50.40 & 46.02 & 45.98 & 65.17 & 49.18 & 54.97\\

$+$ DAPO &45.95 & 51.81 & 21.68 & 51.20 & 46.65 & 47.50 & \textbf{65.84} & 51.03 & 56.33 \\

$+$ GSPO &47.07 & 49.25 & 22.68 & 50.40 & 45.69 & 46.01 & 64.24 & 48.50 & 54.28 \\ 

$+$ AEPO    &46.36 & 51.78 & 23.47 & 50.40 & 46.95 & 45.71 & 64.66 & 50.13 & 55.20 \\

$+$ \textbf{AT\textsuperscript{2}PO (Ours)} & \textbf{49.44} & \textbf{52.99} & \textbf{24.80} & \textbf{56.80} & \textbf{48.81} &\textbf{47.90}&65.32&\textbf{51.81}&\textbf{56.44} \\

\hline 

\rowcolor{pink!20}\multicolumn{10}{c}{\textbf{Backbone Model: Qwen3-8B}} \\

ReAct & 20.66 & 19.05 & 9.56 & 37.60 & 18.66 & 21.16 & 41.81 & 27.37 & 32.19 \\
$+$ GRPO &47.01 &53.69 &21.35 &54.40 &48.03 & 45.70 & 67.42 & 50.17 & 56.29 \\

$+$ DAPO & 49.64 & 53.91 & 24.05 & 56.00 & 49.40 & \textbf{51.99} & 69.02 & 51.90 & 58.53 \\

$+$ GSPO &49.59 &52.55 &24.35 &54.4 &48.56 & 45.56 & 67.75 & 49.66 & 56.15\\

$+$ AEPO &49.17  &52.97  &24.01  &54.40  &48.62  & 49.92 & 68.31 & 51.77 & 57.94 \\

$+$ \textbf{AT\textsuperscript{2}PO (Ours)}&\textbf{51.37} &\textbf{53.97} &\textbf{26.51} &\textbf{56.00} &\textbf{50.15} &51.33 &\textbf{69.51} &\textbf{52.26} &\textbf{58.82}  \\

\hline 

\rowcolor{pink!20}\multicolumn{10}{c}{\textbf{Backbone Model: Qwen2.5-7B}} \\

ReAct  &2.85 & 1.94 & 0.58 & 4.00 & 2.10 & 4.34 & 10.67 & 9.32 & 9.23 \\
$+$ GRPO  &47.94 & 46.89 & 21.27 & 47.20 & 44.48 & 45.56 & 64.86 & 49.92 & 55.20 \\

$+$ DAPO &47.50 & 47.93 & 21.27 & 44.00 & 44.91 & 52.24 & \textbf{65.00} & 50.01 & 56.08 \\

$+$ GSPO  &47.35 & 47.30 & 20.32 & 44.00 & 44.40 & 49.64 & 62.87 & 49.75 & 54.81 \\
$+$ Tree-GRPO & 42.39 & 42.01 & 20.15 & 42.40 & 39.79 & 47.56 & 62.69 & 44.75 & 52.04 \\

$+$ AEPO & 47.05 & 47.53 & 21.03 & 44.00 & 44.51 & 49.00 & 64.13 & 50.21 & 55.45 \\

$+$ \textbf{AT\textsuperscript{2}PO (Ours)}&\textbf{49.58}&\textbf{48.04}&\textbf{22.56}&\textbf{51.20}&\textbf{45.83}&\textbf{52.91}&64.90&\textbf{50.44}&\textbf{56.34} \\

\bottomrule

\end{tabular}
\end{center}
\vspace{-3mm}
\caption{Experiment results on three backbone models across seven datasets. The \textbf{bolded} values indicate the best result in comparisons. Our AT\textsuperscript{2}PO outperforms existing methods in the majority of cases.}
\label{exp:main exp}

\captionsetup{type=figure}
	\centering
\includegraphics[width=1\linewidth]{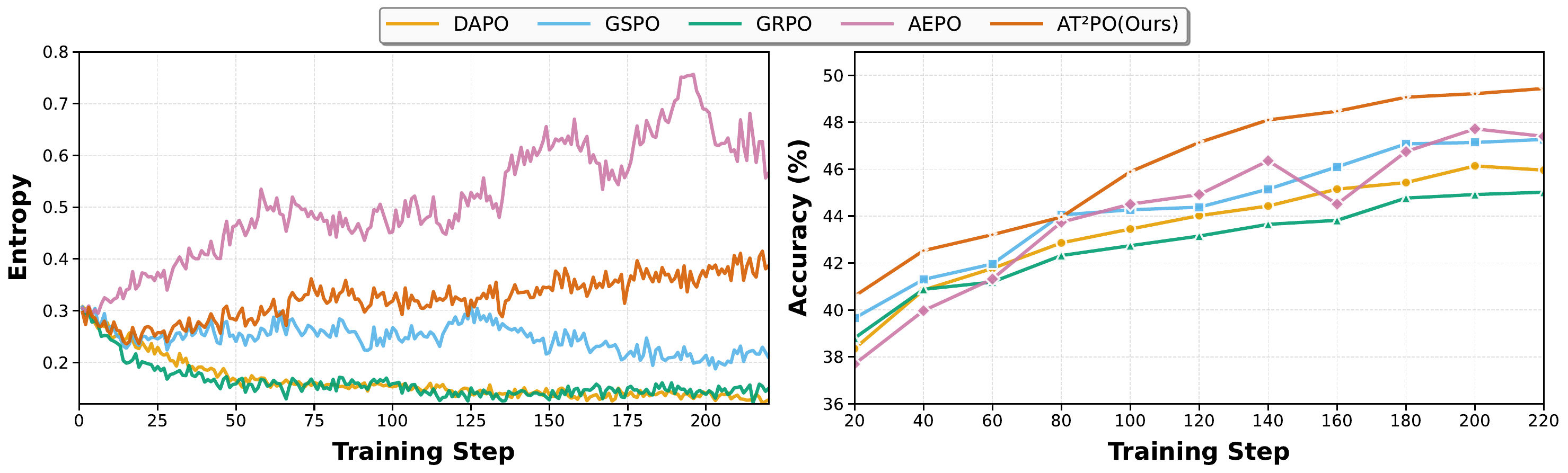}
    \vspace{-4mm}
	\caption{\textit{Left}: The entropy comparison of different methods during training on multihop benchmark. \textit{Right}: Performance comparison between all methods on HotpotQA dataset. Both are based on Qwen3-4B.}
	\label{fig:hotpotqa_exp}
    \vspace{-3mm}
\end{figure*}

\subsection{Experiment Settings}
\textbf{Datasets.}
We evaluate our approach in a tool-integrated knowledge search setting. Following the instructions in Search-R1~\cite{jin2025search}, we implement a lightweight search engine and assess performance on seven widely adopted question answering benchmarks. These datasets are grouped into two categories:
\textbf{Multi-Hop QA}—HotpotQA~\cite{yang2018hotpotqa}, 2WikiMultihopQA~\cite{ho2020constructing}, MuSiQue~\cite{trivedi2022musique}, and Bamboogle~\cite{press2023measuring}; and
\textbf{Single-Hop QA}—Natural Questions (NQ)~\cite{kwiatkowski2019natural}, TriviaQA~\cite{joshi2017triviaqa}, and PopQA~\cite{mallen2023llm_memorization}. All benchmarks are evaluated using Qwen3-4B, Qwen3-8B and Qwen2.5-7B models as the backbone, respectively. We use \textbf{Exact Match (EM)} as the primary evaluation metric across all datasets. 
Critically, our selection avoids reliance on proprietary or costly APIs and imposes minimal hardware requirements, ensuring that our evaluation is both reproducible and focused squarely on algorithmic advances. 

\noindent \textbf{Baselines.} We compare AT\textsuperscript{2}PO against the following widely-accepted RLVR baselines: GRPO \cite{guo2025deepseek}, DAPO \cite{yu2025dapo}, GSPO \cite{zheng2025groupsequencepolicyoptimization}. We also compare AT\textsuperscript{2}PO against two most recent agentic RL baselines: AEPO \cite{dong2025agenticaepo} and Tree-GRPO \cite{ji2025treesearchllmagent}. Note that we notice the performance of Tree-GRPO is highly unstable, even collapses at an early stage on Qwen-3 models. We provide further analysis on the possible cause in the Appendix~\ref{Sec:infra}. We also report the base results by directly inferring via ReAct~\cite{yao2023react}.

\noindent \textbf{Implementation Details.} All experiments are conducted via respective RL recipes without any additional SFT phase. We use a training batch size of $64$, a mini-batch size of $8$ and the maximum response length of $6192$. During rollout, we configure AT\textsuperscript{2}PO with $M=10, L=2$ and $K=6$, while other baselines except DAPO use a global rollout size of $16$. These settings ensure a comparable total token budget across methods. The maximum tool usage is set to $6$. The clipping thresholds for the ATPO objective are set to 3e-3 and 4e-3. All experiments are trained for 240 steps, and we compute the average accuracy across all evaluation samples. We select and report results from the checkpoint with the highest average score. Additional implementation specifics are provided in Section~\ref{sec:implementation}. 

\subsection{Main Results of AT\textsuperscript{2}PO}



\begin{figure}[h]
	\centering
\includegraphics[width=0.47\textwidth]{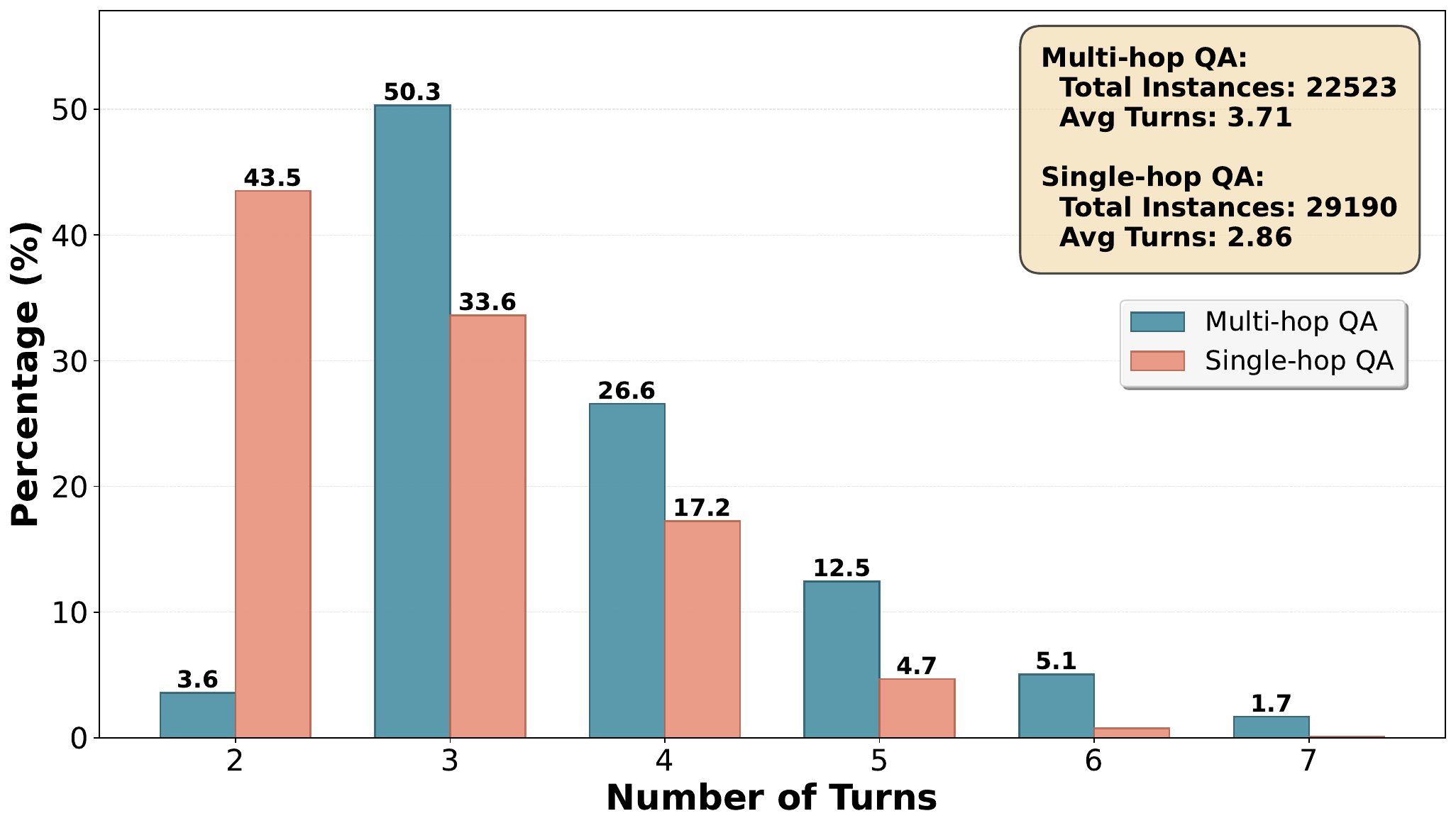}
	\vspace{-2mm}
    \caption{Turn count distribution per sample in the validation set across Multi-hop and Single-hop benchmarks with Qwen3-4B fine-tuned using AT\textsuperscript{2}PO.}
    \label{fig:turn_dist}
    
\end{figure}

We report the performance of all baseline methods and AT\textsuperscript{2}PO across different backbone models and benchmarks in Table~\ref{exp:main exp}. 

Overall, AT\textsuperscript{2}PO achieves the best results among most compared methods, yielding improvements of up to 1.84 percentage points in average over the state-of-the-art baseline across multi-hop and single-hop benchmarks. These results demonstrate the effectiveness and robustness of AT\textsuperscript{2}PO in agentic RL. We note that AT\textsuperscript{2}PO is slightly outperformed on certain single-hop QA datasets. A key reason is that DAPO employs a dynamic sampling strategy that utilizes up to three times the rollout budget compared to all other methods, including AT\textsuperscript{2}PO. When evaluated under an equal computational budget to other baselines, AT\textsuperscript{2}PO achieves the highest average performance across all settings.

Notably, AT\textsuperscript{2}PO exhibits larger gains on multi-hop benchmarks than on single-hop ones. This stems from the fact that multi-hop tasks require significantly more interaction turns to retrieve the correct answer, allowing the turn-level design of AT\textsuperscript{2}PO to fully demonstrate its advantage. As shown in the turn count distribution per sample in the validation set in Figure~\ref{fig:turn_dist}, while 43.5\% of samples in the single-hop benchmark require only a single tool call followed immediately by the answering turn, nearly all samples in the multi-hop benchmark necessitate at least two tool interactions. Consequently, AT\textsuperscript{2}PO achieves state-of-the-art performance across all multi-hop settings, while still delivering strong results on single-hop benchmarks.

Furthermore, we analyze the training dynamics of AT\textsuperscript{2}PO in detail. Figure~\ref{fig:hotpotqa_exp} shows entropy evolution and validation accuracy across training steps. We observe that standard token-level optimization methods such as GRPO suffer from early entropy collapse, which severely limits their ability to explore high-quality trajectories. AEPO, which adapts a modified objective from GPPO~\cite{su2025klear} using a soft clipping strategy to preserve gradients for all tokens, avoids early collapse but exhibits gradual entropy divergence over long-horizon training. In contrast, AT\textsuperscript{2}PO maintains the most stable entropy trajectory, effectively balancing exploration with gradient optimization stability.


\subsection{Analysis}

\subsubsection{Ablation Study on AT\textsuperscript{2}PO}

Table~\ref{tab:ablation} and Figure~\ref{fig:ablation} present an ablation study on Qwen3-4B to assess the contribution of each component in AT\textsuperscript{2}PO, following a progressive modular where modules are incrementally added. This setup enables us to isolate the effect of each design choice under a fixed backbone and consistent task suite.

Starting from a base configuration of standard GRPO with random tree search, replacing it with ATPO yields substantial gains across all multi-hop benchmarks. This highlights that the turn-wise importance sampling and clipping provides a significantly finer-grained optimization objective than vanilla GRPO in multi-turn agentic RL. Next, integrating Entropy-Guided Tree Expansion leads to a further clear performance improvement. This indicates that entropy-aware branching facilitates more effective exploration during tree-based rollouts, enabling the agent to allocate its limited search budget toward more informative decision points. Finally, adding Turn-wise Credit Assignment delivers additional gains and achieves the best overall performance. This demonstrates that providing fine-grained, step-level learning signals to individual turns effectively mitigates the credit assignment problem inherent in sparse-reward agentic tasks.

\begin{table}[t]
    \small
    \centering
    \resizebox{0.47\textwidth}{!}{
    \begin{tabular}{cc}
        \toprule
        \textbf{Model} & \textbf{Avg.} \\
        \midrule
        \textbf{Qwen3-4B with ReAct} & 30.01\\
        \midrule
        Random Tree Search with GRPO Loss & 45.42 \\
        Random Tree Search with ATPO Loss & 47.75 \\
        + Entropy Guided Tree Expansion & 48.33 \\
        + Turn-wise Credit Assignment (\textbf{AT\textsuperscript{2}PO}) & \textbf{48.81} \\
        \bottomrule
    \end{tabular}
    }
    \caption{Ablation Study on progressive molular designs applied to AT\textsuperscript{2}PO on multi-hop benchmarks.}
    \label{tab:ablation}
    \vspace{-4mm}
\end{table}

\subsubsection{Scheme Analysis of Turn-wise Credit Assignment}
\label{sec:ablation-credit}
Table~\ref{tab:stepwise} and Table~\ref{tab:stepwise2} compare different formulations of turn-wise credit assignment by varying both the advantage definition and the value aggregation strategy.
We consider three strategies for estimating the node value $V_n$: (a) child-weighted propagation as defined in in Eq~(3), (b) mean of all child values and (c) mean of all leaf values following TreeRL~\cite{hou2025treerl}. As for advantage computation $f(\cdot)$, our main experiments adopt the simplest formulation $A_n=V_n$. We further compare this with several difference-based alternatives: (a) the TD advantage $A_n=L_n=V_n-V_{n_{\text{parent}}}$, (b) the global advantage $A_n=G_n=V_n-V_{n_{\text{root}}}$ and (c) the combined signal $A_n=L_n+G_n$ as used in prior work~\cite{hou2025treerl}. 

Overall, directly setting $A_n=V_n$ yields best performance across all datasets. Under this formulation, child-weighted aggregation achieves the best results, while child-mean and leaf-mean aggregation remain competitive but slightly inferior. In contrast, difference-based formulations lead to noticeably lower performance. Surprisingly, the combined signal, despite its demonstrated effectiveness in conventional RLVR settings~\cite{yao2023tree}, performs worst among all variants. In the agentic RL, directly using $V_n$ as the advantage provides a cleaner and more compatible learning signal.

\begin{table}[t]
    \small
    \centering
    \resizebox{0.47\textwidth}{!}{
    \begin{tabular}{ccc}
        \toprule
        \textbf{Advantage $A(n)$} & \textbf{Value $V(n)$} & \textbf{Avg.} \\
        \midrule
         $V_n$ & Child Weighted & \textbf{48.81} \\
         $V_n$ & Child Mean & 48.43 \\
         $V_n$ & Leaf Mean & 48.52 \\
         \midrule
         $L_n=V_n-V_{n_{\text{parent}}}$ & Child Weighted & 45.80 \\
         $L_n=V_n-V_{n_{\text{parent}}}$ & Leaf Mean &45.61  \\
         \midrule
         $G_n=V_n-V_{n_{\text{root}}}$ & Child Weighted & 48.19 \\
         $G_n=V_n-V_{n_{\text{root}}}$ & Leaf Mean &46.98  \\
         \midrule
         $L_n+G_n$ & Leaf Mean & 44.18  \\
        \bottomrule
    \end{tabular}
    }
    \caption{Ablation on the effectiveness of different turn-wise credit assignment choices.}
    \label{tab:stepwise}
    \vspace{-4mm}
\end{table}

\section{Conclusion}
In this paper, we present AT\textsuperscript{2}PO, a unified framework for multi-turn agentic RL. Through Entropy-Guided Tree Expansion, we enable strategic exploration at uncertain decision points; via Turn-wise Credit Assignment, we propagate sparse outcome rewards to derive fine-grained supervision signals; and with Agentic Turn-based Policy Optimization, we align policy updates with the natural turn-based structure. Extensive experiments across seven benchmarks demonstrate consistent improvements over strong baselines, validating the effectiveness of each component. Our framework offers a promising approach to training LLM agents for complex multi-turn reasoning and tool-use tasks.

\section{Limitations}
While AT\textsuperscript{2}PO achieves strong empirical performance, its tree-based expansion introduces additional computational overhead compared to linear rollouts particularly when sufficient computational resources are available, as the expansion process requires multiple sequential iterations. Future work will explore more efficient rollout strategies through enhanced parallelization. Additionally, we aim to evaluate AT\textsuperscript{2}PO in a broader range of agentic environments to further assess its generalization and robustness.

\bibliography{custom}

@inproceedings{
yao2023react,
title={ReAct: Synergizing Reasoning and Acting in Language Models},
author={Shunyu Yao and Jeffrey Zhao and Dian Yu and Nan Du and Izhak Shafran and Karthik R Narasimhan and Yuan Cao},
booktitle={The Eleventh International Conference on Learning Representations },
year={2023},
url={https://openreview.net/forum?id=WE_vluYUL-X}
}

@inproceedings{
schick2023toolformer,
title={Toolformer: Language Models Can Teach Themselves to Use Tools},
author={Timo Schick and Jane Dwivedi-Yu and Roberto Dessi and Roberta Raileanu and Maria Lomeli and Eric Hambro and Luke Zettlemoyer and Nicola Cancedda and Thomas Scialom},
booktitle={Thirty-seventh Conference on Neural Information Processing Systems},
year={2023},
url={https://openreview.net/forum?id=Yacmpz84TH}
}

@inproceedings{shen2024smartcal,
  title={SMARTCAL: An Approach to Self-Aware Tool-Use Evaluation and Calibration},
  author={Shen, Yuanhao and Zhu, Xiaodan and Chen, Lei},
  booktitle={Proceedings of the 2024 Conference on Empirical Methods in Natural Language Processing: Industry Track},
  pages={774--789},
  year={2024}
}

@article{guo2025deepseek,
  title={Deepseek-r1: Incentivizing reasoning capability in llms via reinforcement learning},
  author={Guo, Daya and Yang, Dejian and Zhang, Haowei and Song, Junxiao and Zhang, Ruoyu and Xu, Runxin and Zhu, Qihao and Ma, Shirong and Wang, Peiyi and Bi, Xiao and others},
  journal={arXiv preprint arXiv:2501.12948},
  year={2025}
}

@article{jin2025search,
  title={Search-r1: Training llms to reason and leverage search engines with reinforcement learning},
  author={Jin, Bowen and Zeng, Hansi and Yue, Zhenrui and Yoon, Jinsung and Arik, Sercan and Wang, Dong and Zamani, Hamed and Han, Jiawei},
  journal={arXiv preprint arXiv:2503.09516},
  year={2025}
}

@article{wang2025ragen,
  title={Ragen: Understanding self-evolution in llm agents via multi-turn reinforcement learning},
  author={Wang, Zihan and Wang, Kangrui and Wang, Qineng and Zhang, Pingyue and Li, Linjie and Yang, Zhengyuan and Jin, Xing and Yu, Kefan and Nguyen, Minh Nhat and Liu, Licheng and others},
  journal={arXiv preprint arXiv:2504.20073},
  year={2025}
}

@article{wu2025webdancer,
  title={Webdancer: Towards autonomous information seeking agency},
  author={Wu, Jialong and Li, Baixuan and Fang, Runnan and Yin, Wenbiao and Zhang, Liwen and Tao, Zhengwei and Zhang, Dingchu and Xi, Zekun and Fu, Gang and Jiang, Yong and others},
  journal={arXiv preprint arXiv:2505.22648},
  year={2025}
}

@article{li2025websailor,
  title={WebSailor: Navigating Super-human Reasoning for Web Agent},
  author={Li, Kuan and Zhang, Zhongwang and Yin, Huifeng and Zhang, Liwen and Ou, Litu and Wu, Jialong and Yin, Wenbiao and Li, Baixuan and Tao, Zhengwei and Wang, Xinyu and others},
  journal={arXiv preprint arXiv:2507.02592},
  year={2025}
}

@article{dong2025agenticarpo,
  title={Agentic reinforced policy optimization},
  author={Dong, Guanting and Mao, Hangyu and Ma, Kai and Bao, Licheng and Chen, Yifei and Wang, Zhongyuan and Chen, Zhongxia and Du, Jiazhen and Wang, Huiyang and Zhang, Fuzheng and others},
  journal={arXiv preprint arXiv:2507.19849},
  year={2025}
}

@article{dong2025agenticaepo,
  title={Agentic entropy-balanced policy optimization},
  author={Dong, Guanting and Bao, Licheng and Wang, Zhongyuan and Zhao, Kangzhi and Li, Xiaoxi and Jin, Jiajie and Yang, Jinghan and Mao, Hangyu and Zhang, Fuzheng and Gai, Kun and others},
  journal={arXiv preprint arXiv:2510.14545},
  year={2025}
}

@article{feng2025group,
  title={Group-in-group policy optimization for llm agent training},
  author={Feng, Lang and Xue, Zhenghai and Liu, Tingcong and An, Bo},
  journal={arXiv preprint arXiv:2505.10978},
  year={2025}
}

@article{yan2025memory,
  title={Memory-r1: Enhancing large language model agents to manage and utilize memories via reinforcement learning},
  author={Yan, Sikuan and Yang, Xiufeng and Huang, Zuchao and Nie, Ercong and Ding, Zifeng and Li, Zonggen and Ma, Xiaowen and Kersting, Kristian and Pan, Jeff Z and Sch{\"u}tze, Hinrich and others},
  journal={arXiv preprint arXiv:2508.19828},
  year={2025}
}

@article{yu2025memagent,
  title={MemAgent: Reshaping Long-Context LLM with Multi-Conv RL-based Memory Agent},
  author={Yu, Hongli and Chen, Tinghong and Feng, Jiangtao and Chen, Jiangjie and Dai, Weinan and Yu, Qiying and Zhang, Ya-Qin and Ma, Wei-Ying and Liu, Jingjing and Wang, Mingxuan and others},
  journal={arXiv preprint arXiv:2507.02259},
  year={2025}
}

@article{wu2025mmsearch,
  title={MMSearch-R1: Incentivizing LMMs to Search},
  author={Wu, Jinming and Deng, Zihao and Li, Wei and Liu, Yiding and You, Bo and Li, Bo and Ma, Zejun and Liu, Ziwei},
  journal={arXiv preprint arXiv:2506.20670},
  year={2025}
}

@article{geng2025webwatcher,
  title={Webwatcher: Breaking new frontier of vision-language deep research agent},
  author={Geng, Xinyu and Xia, Peng and Zhang, Zhen and Wang, Xinyu and Wang, Qiuchen and Ding, Ruixue and Wang, Chenxi and Wu, Jialong and Zhao, Yida and Li, Kuan and others},
  journal={arXiv preprint arXiv:2508.05748},
  year={2025}
}

@article{wang2025otc,
  title={Otc: Optimal tool calls via reinforcement learning},
  author={Wang, Hongru and Qian, Cheng and Zhong, Wanjun and Chen, Xiusi and Qiu, Jiahao and Huang, Shijue and Jin, Bowen and Wang, Mengdi and Wong, Kam-Fai and Ji, Heng},
  journal={arXiv e-prints},
  pages={arXiv--2504},
  year={2025}
}

@article{yao2023tree,
  title={Tree of thoughts: Deliberate problem solving with large language models},
  author={Yao, Shunyu and Yu, Dian and Zhao, Jeffrey and Shafran, Izhak and Griffiths, Tom and Cao, Yuan and Narasimhan, Karthik},
  journal={Advances in neural information processing systems},
  volume={36},
  pages={11809--11822},
  year={2023}
}

@misc{ji2025treesearchllmagent,
      title={Tree Search for LLM Agent Reinforcement Learning}, 
      author={Yuxiang Ji and Ziyu Ma and Yong Wang and Guanhua Chen and Xiangxiang Chu and Liaoni Wu},
      year={2025},
      eprint={2509.21240},
      archivePrefix={arXiv},
      primaryClass={cs.LG},
      url={https://arxiv.org/abs/2509.21240}, 
}

@misc{hou2025treerl,
      title={TreeRL: LLM Reinforcement Learning with On-Policy Tree Search}, 
      author={Zhenyu Hou and Ziniu Hu and Yujiang Li and Rui Lu and Jie Tang and Yuxiao Dong},
      year={2025},
      eprint={2506.11902},
      archivePrefix={arXiv},
      primaryClass={cs.LG},
      url={https://arxiv.org/abs/2506.11902}, 
}

@misc{shao2024deepseekmathpushinglimitsmathematical,
      title={DeepSeekMath: Pushing the Limits of Mathematical Reasoning in Open Language Models}, 
      author={Zhihong Shao and Peiyi Wang and Qihao Zhu and Runxin Xu and Junxiao Song and Xiao Bi and Haowei Zhang and Mingchuan Zhang and Y. K. Li and Y. Wu and Daya Guo},
      year={2024},
      eprint={2402.03300},
      archivePrefix={arXiv},
      primaryClass={cs.CL},
      url={https://arxiv.org/abs/2402.03300}, 
}

@misc{schulman2017proximal,
      title={Proximal Policy Optimization Algorithms}, 
      author={John Schulman and Filip Wolski and Prafulla Dhariwal and Alec Radford and Oleg Klimov},
      year={2017},
      eprint={1707.06347},
      archivePrefix={arXiv},
      primaryClass={cs.LG},
      url={https://arxiv.org/abs/1707.06347}, 
}

@inproceedings{yang2018hotpotqa,
    title = "{H}otpot{QA}: A Dataset for Diverse, Explainable Multi-hop Question Answering",
    author = "Yang, Zhilin  and
      Qi, Peng  and
      Zhang, Saizheng  and
      Bengio, Yoshua  and
      Cohen, William  and
      Salakhutdinov, Ruslan  and
      Manning, Christopher D.",
    editor = "Riloff, Ellen  and
      Chiang, David  and
      Hockenmaier, Julia  and
      Tsujii, Jun{'}ichi",
    booktitle = "Proceedings of the 2018 Conference on Empirical Methods in Natural Language Processing",
    month = oct # "-" # nov,
    year = "2018",
    address = "Brussels, Belgium",
    publisher = "Association for Computational Linguistics",
    url = "https://aclanthology.org/D18-1259/",
    doi = "10.18653/v1/D18-1259",
    pages = "2369--2380"
}

@inproceedings{ho2020constructing,
    title = "Constructing A Multi-hop {QA} Dataset for Comprehensive Evaluation of Reasoning Steps",
    author = "Ho, Xanh  and
      Duong Nguyen, Anh-Khoa  and
      Sugawara, Saku  and
      Aizawa, Akiko",
    editor = "Scott, Donia  and
      Bel, Nuria  and
      Zong, Chengqing",
    booktitle = "Proceedings of the 28th International Conference on Computational Linguistics",
    month = dec,
    year = "2020",
    address = "Barcelona, Spain (Online)",
    publisher = "International Committee on Computational Linguistics",
    url = "https://aclanthology.org/2020.coling-main.580/",
    doi = "10.18653/v1/2020.coling-main.580",
    pages = "6609--6625"
}

@article{trivedi2022musique,
    title = "{M}u{S}i{Q}ue: Multihop Questions via Single-hop Question Composition",
    author = "Trivedi, Harsh  and
      Balasubramanian, Niranjan  and
      Khot, Tushar  and
      Sabharwal, Ashish",
    editor = "Roark, Brian  and
      Nenkova, Ani",
    journal = "Transactions of the Association for Computational Linguistics",
    volume = "10",
    year = "2022",
    address = "Cambridge, MA",
    publisher = "MIT Press",
    url = "https://aclanthology.org/2022.tacl-1.31/",
    doi = "10.1162/tacl_a_00475",
    pages = "539--554"
}

@misc{press2023measuring,
      title={Measuring and Narrowing the Compositionality Gap in Language Models}, 
      author={Ofir Press and Muru Zhang and Sewon Min and Ludwig Schmidt and Noah A. Smith and Mike Lewis},
      year={2023},
      eprint={2210.03350},
      archivePrefix={arXiv},
      primaryClass={cs.CL},
      url={https://arxiv.org/abs/2210.03350}, 
}

@article{kwiatkowski2019natural,
    title = "Natural Questions: A Benchmark for Question Answering Research",
    author = "Kwiatkowski, Tom  and
      Palomaki, Jennimaria  and
      Redfield, Olivia  and
      Collins, Michael  and
      Parikh, Ankur  and
      Alberti, Chris  and
      Epstein, Danielle  and
      Polosukhin, Illia  and
      Devlin, Jacob  and
      Lee, Kenton  and
      Toutanova, Kristina  and
      Jones, Llion  and
      Kelcey, Matthew  and
      Chang, Ming-Wei  and
      Dai, Andrew M.  and
      Uszkoreit, Jakob  and
      Le, Quoc  and
      Petrov, Slav",
    editor = "Lee, Lillian  and
      Johnson, Mark  and
      Roark, Brian  and
      Nenkova, Ani",
    journal = "Transactions of the Association for Computational Linguistics",
    volume = "7",
    year = "2019",
    address = "Cambridge, MA",
    publisher = "MIT Press",
    url = "https://aclanthology.org/Q19-1026/",
    doi = "10.1162/tacl_a_00276",
    pages = "452--466"
}

@inproceedings{joshi2017triviaqa,
    title = "{T}rivia{QA}: A Large Scale Distantly Supervised Challenge Dataset for Reading Comprehension",
    author = "Joshi, Mandar  and
      Choi, Eunsol  and
      Weld, Daniel  and
      Zettlemoyer, Luke",
    editor = "Barzilay, Regina  and
      Kan, Min-Yen",
    booktitle = "Proceedings of the 55th Annual Meeting of the Association for Computational Linguistics (Volume 1: Long Papers)",
    month = jul,
    year = "2017",
    address = "Vancouver, Canada",
    publisher = "Association for Computational Linguistics",
    url = "https://aclanthology.org/P17-1147/",
    doi = "10.18653/v1/P17-1147",
    pages = "1601--1611"
}

@article{mallen2023llm_memorization,
  title={When Not to Trust Language Models: Investigating Effectiveness and Limitations of Parametric and Non-Parametric Memories },
  author={ Mallen, Alex and Asai,Akari and  Zhong, Victor and Das, Rajarshi and Hajishirzi, Hannaneh and Khashabi, Daniel},
  journal={ arXiv preprint },
  year={ 2022 }
}

@inproceedings{yu2025dapo,
title={{DAPO}: An Open-Source {LLM} Reinforcement Learning System at Scale},
author={Qiying Yu and Zheng Zhang and Ruofei Zhu and Yufeng Yuan and Xiaochen Zuo and YuYue and Weinan Dai and Tiantian Fan and Gaohong Liu and Juncai Liu and LingJun Liu and Xin Liu and Haibin Lin and Zhiqi Lin and Bole Ma and Guangming Sheng and Yuxuan Tong and Chi Zhang and Mofan Zhang and Ru Zhang and Wang Zhang and Hang Zhu and Jinhua Zhu and Jiaze Chen and Jiangjie Chen and Chengyi Wang and Hongli Yu and Yuxuan Song and Xiangpeng Wei and Hao Zhou and Jingjing Liu and Wei-Ying Ma and Ya-Qin Zhang and Lin Yan and Yonghui Wu and Mingxuan Wang},
booktitle={The Thirty-ninth Annual Conference on Neural Information Processing Systems},
year={2025},
url={https://openreview.net/forum?id=2a36EMSSTp}
}

@misc{zheng2025groupsequencepolicyoptimization,
      title={Group Sequence Policy Optimization}, 
      author={Chujie Zheng and Shixuan Liu and Mingze Li and Xiong-Hui Chen and Bowen Yu and Chang Gao and Kai Dang and Yuqiong Liu and Rui Men and An Yang and Jingren Zhou and Junyang Lin},
      year={2025},
      eprint={2507.18071},
      archivePrefix={arXiv},
      primaryClass={cs.LG},
      url={https://arxiv.org/abs/2507.18071}, 
}

@misc{yang2025qwen3technicalreport,
      title={Qwen3 Technical Report}, 
      author={An Yang and Anfeng Li and Baosong Yang and Beichen Zhang and Binyuan Hui and Bo Zheng and Bowen Yu and Chang Gao and Chengen Huang and Chenxu Lv and Chujie Zheng and Dayiheng Liu and Fan Zhou and Fei Huang and Feng Hu and Hao Ge and Haoran Wei and Huan Lin and Jialong Tang and Jian Yang and Jianhong Tu and Jianwei Zhang and Jianxin Yang and Jiaxi Yang and Jing Zhou and Jingren Zhou and Junyang Lin and Kai Dang and Keqin Bao and Kexin Yang and Le Yu and Lianghao Deng and Mei Li and Mingfeng Xue and Mingze Li and Pei Zhang and Peng Wang and Qin Zhu and Rui Men and Ruize Gao and Shixuan Liu and Shuang Luo and Tianhao Li and Tianyi Tang and Wenbiao Yin and Xingzhang Ren and Xinyu Wang and Xinyu Zhang and Xuancheng Ren and Yang Fan and Yang Su and Yichang Zhang and Yinger Zhang and Yu Wan and Yuqiong Liu and Zekun Wang and Zeyu Cui and Zhenru Zhang and Zhipeng Zhou and Zihan Qiu},
      year={2025},
      eprint={2505.09388},
      archivePrefix={arXiv},
      primaryClass={cs.CL},
      url={https://arxiv.org/abs/2505.09388}, 
}

@article{wang2022text,
  title={Text Embeddings by Weakly-Supervised Contrastive Pre-training},
  author={Wang, Liang and Yang, Nan and Huang, Xiaolong and Jiao, Binxing and Yang, Linjun and Jiang, Daxin and Majumder, Rangan and Wei, Furu},
  journal={arXiv preprint arXiv:2212.03533},
  year={2022}
}

@article{christiano2017deep,
  title={Deep reinforcement learning from human preferences},
  author={Christiano, Paul F and Leike, Jan and Brown, Tom and Martic, Miljan and Legg, Shane and Amodei, Dario},
  journal={Advances in neural information processing systems},
  volume={30},
  year={2017}
}

@article{stiennon2020learning,
  title={Learning to summarize with human feedback},
  author={Stiennon, Nisan and Ouyang, Long and Wu, Jeffrey and Ziegler, Daniel and Lowe, Ryan and Voss, Chelsea and Radford, Alec and Amodei, Dario and Christiano, Paul F},
  journal={Advances in neural information processing systems},
  volume={33},
  pages={3008--3021},
  year={2020}
}

@misc{lambert2025tulu3pushingfrontiers,
      title={Tulu 3: Pushing Frontiers in Open Language Model Post-Training}, 
      author={Nathan Lambert and Jacob Morrison and Valentina Pyatkin and Shengyi Huang and Hamish Ivison and Faeze Brahman and Lester James V. Miranda and Alisa Liu and Nouha Dziri and Shane Lyu and Yuling Gu and Saumya Malik and Victoria Graf and Jena D. Hwang and Jiangjiang Yang and Ronan Le Bras and Oyvind Tafjord and Chris Wilhelm and Luca Soldaini and Noah A. Smith and Yizhong Wang and Pradeep Dasigi and Hannaneh Hajishirzi},
      year={2025},
      eprint={2411.15124},
      archivePrefix={arXiv},
      primaryClass={cs.CL},
      url={https://arxiv.org/abs/2411.15124}, 
}

@misc{wen2025reinforcementlearningverifiablerewards,
      title={Reinforcement Learning with Verifiable Rewards Implicitly Incentivizes Correct Reasoning in Base LLMs}, 
      author={Xumeng Wen and Zihan Liu and Shun Zheng and Shengyu Ye and Zhirong Wu and Yang Wang and Zhijian Xu and Xiao Liang and Junjie Li and Ziming Miao and Jiang Bian and Mao Yang},
      year={2025},
      eprint={2506.14245},
      archivePrefix={arXiv},
      primaryClass={cs.AI},
      url={https://arxiv.org/abs/2506.14245}, 
}

@misc{wang2025reinforcementlearningreasoninglarge,
      title={Reinforcement Learning for Reasoning in Large Language Models with One Training Example}, 
      author={Yiping Wang and Qing Yang and Zhiyuan Zeng and Liliang Ren and Liyuan Liu and Baolin Peng and Hao Cheng and Xuehai He and Kuan Wang and Jianfeng Gao and Weizhu Chen and Shuohang Wang and Simon Shaolei Du and Yelong Shen},
      year={2025},
      eprint={2504.20571},
      archivePrefix={arXiv},
      primaryClass={cs.LG},
      url={https://arxiv.org/abs/2504.20571}, 
}

@article{chen2025minimax,
  title={MiniMax-M1: Scaling Test-Time Compute Efficiently with Lightning Attention},
  author={Chen, Aili and Li, Aonian and Gong, Bangwei and Jiang, Binyang and Fei, Bo and Yang, Bo and Shan, Boji and Yu, Changqing and Wang, Chao and Zhu, Cheng and others},
  journal={arXiv preprint arXiv:2506.13585},
  year={2025}
}

@article{wang2025aspo,
  title={Aspo: Asymmetric importance sampling policy optimization},
  author={Wang, Jiakang and Liu, Runze and Lin, Lei and Hu, Wenping and Li, Xiu and Zhang, Fuzheng and Zhou, Guorui and Gai, Kun},
  journal={arXiv preprint arXiv:2510.06062},
  year={2025}
}

@article{su2025gppo,
  title={CE-GPPO: Coordinating Entropy via Gradient-Preserving Clipping Policy Optimization in Reinforcement Learning},
  author={Su, Zhenpeng and Pan, Leiyu and Lv, Minxuan and Li, Yuntao and Hu, Wenping and Zhang, Fuzheng and Gai, Kun and Zhou, Guorui},
  journal={arXiv preprint arXiv:2509.20712},
  year={2025}
}

@misc{gao2025softadaptivepolicyoptimization,
      title={Soft Adaptive Policy Optimization}, 
      author={Chang Gao and Chujie Zheng and Xiong-Hui Chen and Kai Dang and Shixuan Liu and Bowen Yu and An Yang and Shuai Bai and Jingren Zhou and Junyang Lin},
      year={2025},
      eprint={2511.20347},
      archivePrefix={arXiv},
      primaryClass={cs.LG},
      url={https://arxiv.org/abs/2511.20347}, 
}

@article{su2025klear,
  title={Klear-reasoner: Advancing reasoning capability via gradient-preserving clipping policy optimization},
  author={Su, Zhenpeng and Pan, Leiyu and Bai, Xue and Liu, Dening and Dong, Guanting and Huang, Jiaming and Hu, Wenping and Zhang, Fuzheng and Gai, Kun and Zhou, Guorui},
  journal={arXiv preprint arXiv:2508.07629},
  year={2025}
}

@article{zheng2023click,
  title={Click: Controllable text generation with sequence likelihood contrastive learning},
  author={Zheng, Chujie and Ke, Pei and Zhang, Zheng and Huang, Minlie},
  journal={arXiv preprint arXiv:2306.03350},
  year={2023}
}

@misc{vllm2025retokenization,
  title        = {No More Retokenization Drift: Returning Token IDs via the OpenAI Compatible API Matters in Agent RL},
  author       = {{vLLM Team}},
  year         = {2025},
  month        = oct,
  day          = {22},
  howpublished = {\url{https://blog.vllm.ai/2025/10/22/agent-lightning.html}},
  note         = {Accessed: 2026-01-05}
}

@misc{qwen2025qwen25technicalreport,
      title={Qwen2.5 Technical Report}, 
      author={Qwen and : and An Yang and Baosong Yang and Beichen Zhang and Binyuan Hui and Bo Zheng and Bowen Yu and Chengyuan Li and Dayiheng Liu and Fei Huang and Haoran Wei and Huan Lin and Jian Yang and Jianhong Tu and Jianwei Zhang and Jianxin Yang and Jiaxi Yang and Jingren Zhou and Junyang Lin and Kai Dang and Keming Lu and Keqin Bao and Kexin Yang and Le Yu and Mei Li and Mingfeng Xue and Pei Zhang and Qin Zhu and Rui Men and Runji Lin and Tianhao Li and Tianyi Tang and Tingyu Xia and Xingzhang Ren and Xuancheng Ren and Yang Fan and Yang Su and Yichang Zhang and Yu Wan and Yuqiong Liu and Zeyu Cui and Zhenru Zhang and Zihan Qiu},
      year={2025},
      eprint={2412.15115},
      archivePrefix={arXiv},
      primaryClass={cs.CL},
      url={https://arxiv.org/abs/2412.15115}, 
}

@article{sheng2024hybridflow,
  title   = {HybridFlow: A Flexible and Efficient RLHF Framework},
  author  = {Guangming Sheng and Chi Zhang and Zilingfeng Ye and Xibin Wu and Wang Zhang and Ru Zhang and Yanghua Peng and Haibin Lin and Chuan Wu},
  year    = {2024},
  journal = {arXiv preprint arXiv: 2409.19256}
}

@inproceedings{zhang2024framework,
  title={A framework for training large language models for code generation via proximal policy optimization},
  author={Zhang, Chi and Sheng, Guangming and Liu, Siyao and Li, Jiahao and Feng, Ziyuan and Liu, Zherui and Liu, Xin and Jia, Xiaoying and Peng, Yanghua and Lin, Haibin and others},
  booktitle={NL2Code Workshop of ACM KDD},
  year={2024}
}

\clearpage

\appendix

\section{Analysis of ATPO Objective}
\label{sec:formulas}

In this section, we clarify the distinctions among token-level, sequence-level, and turn-level policy optimization objectives, highlighting how each approach leverages different granularities of policy updates.

\textbf{Token-Level Optimization. } GRPO~\cite{shao2024deepseekmathpushinglimitsmathematical} computes the relative advantage of each response within a group and optimizes the following token-level objective:
\begin{equation}
\begin{split}
& \mathcal{J}_{GRPO}(\theta) = \mathbb{E}_{x \sim \mathcal{D},\, \{y_i\}_{i=1}^G \sim \pi_{\theta_{old}}(\cdot|x)} \Bigg[ \\
& \quad \frac{1}{G} \sum_{i=1}^G \frac{1}{|y_i|} \sum_{t=1}^{|y_i|} M_{i,t}\min \Bigg( r_{i,t}(\theta) \widehat{A}_{i,t},\, \\
& \qquad \text{clip}\left(r_{i,t}(\theta), 1-\varepsilon, 1+\varepsilon \Bigg) \widehat{A}_{i,t} \right) \Bigg], \\
\end{split}
\end{equation}
where the importance sampling ratio is defined at the token level as
\begin{equation}
    r_{i,t}(\theta) = \frac{\pi_{\theta}(y_{i,t}|x, y_{i,<t})}{\pi_{\theta_{old}}(y_{i,t}|x, y_{i,<t})}.
\end{equation}
Notably, All tokens in a sample share the same advantage $\hat{A}_{i,t}=\hat{A}_i$.

\textbf{Sequence-Level Optimization.} In contrast, GSPO~\cite{zheng2025groupsequencepolicyoptimization} adopts a sequence-level perspective by defining the importance ratio based on the entire response likelihood:
\begin{equation}
\begin{split}
& \mathcal{J}_{GSPO}(\theta) = \mathbb{E}_{x \sim \mathcal{D}, \{y_i\}_{i=1}^G \sim \pi_{\theta_{old}}(\cdot | x)} \Bigg[ \\
& \quad \frac{1}{G} \sum_{i=1}^G \frac{1}{|y_i|} \sum_{t=1}^{|y_i|} M_{i,t}\min \Bigg( s_{i,t}(\theta) \hat{A}_{i,t}, \\
& \qquad \operatorname{clip} \left( s_{i,t}(\theta), 1 - \varepsilon_l, 1 + \varepsilon_r \right) \hat{A}_{i,t} \Bigg) \Bigg]
\end{split}
\label{eq:gspo}
\end{equation}
with the sequence-aware importance ratio given by
\begin{equation}
\begin{split}
 & s_{i,t}(\theta) = \operatorname{sg} \left[ \Bigg( \frac{\pi_{\theta}(y_i|x)}{\pi_{\theta_{old}}(y_i|x)} \Bigg)^{\frac{1}{|y_i|}} \right] \\
 & \quad \cdot \frac{\pi_\theta(y_{i,t} | x, y_{i,<t})}{\operatorname{sg} \left[ \pi_\theta(y_{i,t} | x, y_{i,<t}) \right]}.
\end{split}
\end{equation}
Here, $sg[\cdot]$ denotes the stop-gradient operator. By anchoring the importance weight to the full-sequence likelihood (following~\cite{zheng2023click}), GSPO ensures that clipping is applied coherently across all tokens in a response, reflecting an all-or-nothing treatment of off-policy sequences.

\textbf{Turn-Level Optimization.} Our proposed Adaptive Turn-level Policy Optimization (ATPO) introduces a turn-aware objective that balances fine-grained control with sequence coherence. The objective is:
\begin{equation}
\begin{split}
& \mathcal{J}_{ATPO}(\theta) = \mathbb{E}_{x \sim \mathcal{D}, \{y_i\}_{i=1}^G \sim \pi_{\theta_{old}}(\cdot | x)} \Bigg[ \\
& \quad \frac{1}{G} \sum_{i=1}^G \frac{1}{|y_i|} \sum_{t=1}^{|y_i|} M_{i,t}\min \Bigg( s_{i,t}^{turn}(\theta) \hat{A}_{i,t}, \\
& \qquad \operatorname{clip} \left( s_{i,t}^{turn}(\theta), 1 - \varepsilon_l, 1 + \varepsilon_r \right) \hat{A}_{i,t} \Bigg) \Bigg],
\end{split}
\label{eq:gspo-turn}
\end{equation}
where the turn-level importance ratio is defined as
\begin{equation}
\begin{split}
 & s_{i,t}^{turn}(\theta) = \frac{\pi_\theta(y_{i,t} | x, y_{i,<t})}{\operatorname{sg} \left[ \pi_\theta(y_{i,t} | x, y_{i,<t}) \right]} \\
 & \quad \cdot \operatorname{sg} \left[ \Bigg( \frac{\pi_{\theta}(y_i^{k(t)}|x, y_i^{<k(t)})}{\pi_{\theta_{old}}(y_i^{k(t)}|x, y_i^{<k(t)})} \Bigg)^{\frac{1}{|y_i^{k(t)}|}} \right].
\end{split}
\label{eq:turn-importance-ratio}
\end{equation}
Here $y_{i,t}$ belongs to the $k(t)$-th response turn $y_i^{k(t)}$. This formulation enables selective gradient propagation: if tokens from a particular turn are highly off-policy, their contribution to the gradient can be suppressed without discarding updates from other more on-policy turns. The gradient is derived as follows,
\begin{equation}
\begin{split}
& \nabla_{\theta}\mathcal{J}_{AT^2PO}(\theta) = \mathbb{E}_{x \sim \mathcal{D}, \{y_i\}_{i=1}^G \sim \pi_{\theta_{old}}(\cdot | x)} \Bigg[ \\
& \quad \frac{1}{G} \sum_{i=1}^G \frac{1}{|y_i|} \sum_{t=1}^{|y_i|} M_{i,t} s_{i,t}^{turn}(\theta) \hat{A}_{i,t} \\
& \qquad  \cdot\nabla_{\theta}\operatorname{log}\pi_{\theta} (y_{i,t}|x,y_{i,<t}) \Bigg]
\end{split}
\label{eq:gspo-turn}
\end{equation}

To better assess the effectiveness of turn-level optimization in ATPO, we also introduce a diagnostic metric called turn entropy $\mathcal{H}_{turn}$, which quantifies the degree of variation in policy updates across different turns within a trajectory:
\begin{equation}
\begin{split}
\mathcal{H}_{turn} = & \frac{1}{B} \sum_{i=1}^{B} \frac{ -\sum_{j=1}^{N_i} p_{i,j} \log(p_{i,j}) }{ \log(N_i) }, \\
 \quad p_{i,j} = & \frac{ \exp\left(\frac{KL_{i,j}}{KL_{i,seq}}\right) }{ \sum_{k=1}^{N_i} \exp\left(\frac{KL_{i,k}}{KL_{i,seq}}\right) },
 \end{split}
\end{equation}
where $N_i$ is number of turns of the $i$-th sample in batch, $KL_{i,j}$ is the KL divergence between the old and new policies at turn $j$, and $KL_{i, seq}$ is the total KL divergence of the full sequence.

We normalize $\mathcal{H}_{turn}$ to $[0,1]$. A special value of $\mathcal{H}_{turn}=1$ indicates uniform policy updates across all turns, while $\mathcal{H}_{turn}=0$ occurs when every sample contains only one turn. Conversely, smaller values signify greater heterogeneity in turn-level updates, highlighting scenarios where turn-aware optimization is particularly beneficial. 

\begin{figure}
    \centering
    \includegraphics[width=0.49\textwidth]{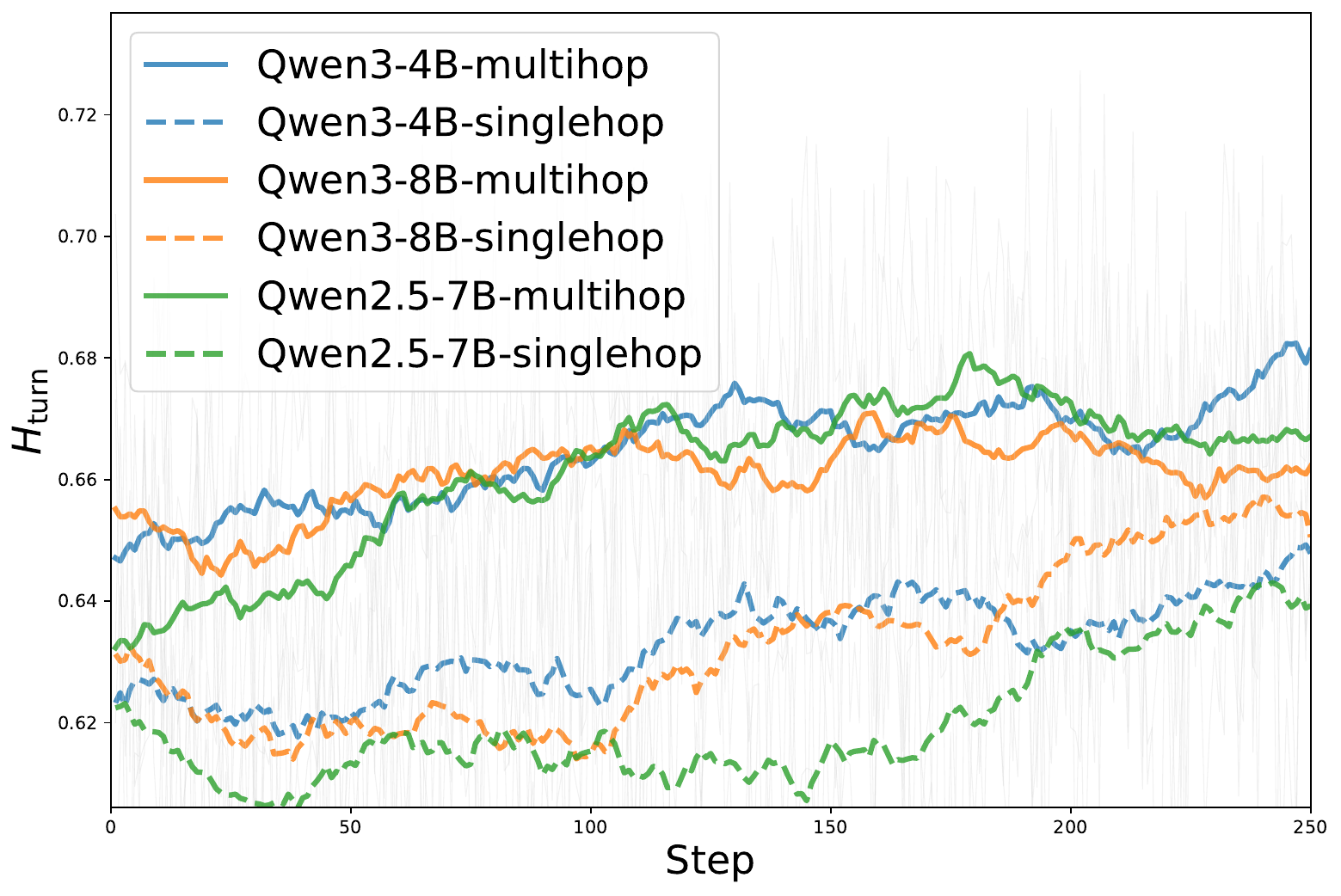}
    \caption{$H_{turn}$ alongside training steps of AT\textsuperscript{2}PO in all experiment settings.}
    \label{fig:Hturn}
\end{figure}

Figure~\ref{fig:Hturn} plots $H_{turn}$ throughout training across all experiment settings. The entropy stabilizes around 0.66 for multi-hop benchmarks and 0.62 for single-hop benchmarks, confirming significant variation in turn-level policy update and justifying the need for turn-granular optimization. Interestingly, we find that $H_{turn}$ gradually increases during training across all settings, suggesting that as the policy converges, update magnitudes become more balanced across turns, which is a sign of stabilized and high-quality learning.

\section{Implementation Details}
\label{sec:implementation}
\subsection{Reward Design}
\label{sec:reward_design}

We adopt a binary rule-based reward function to obtain the final outcome reward in our agentic RL framework, focusing on both answer correctness and format completeness of LLM's output. To be detailed, we follow the reward modeling setting of Search-R1 \cite{jin2025search} which utilizes the EM-score as primary reward. Meanwhile, we also introduce the appropriate format constraint in this process.

\textbf{Exact Match Reward} Let $\hat{y}$ denote the agent’s predicted final answer and $y^{*}$ the ground-truth answer. We leverage an \textbf{Exact Match (EM)} score as the primary metric of answer correctness:
\begin{equation}
r_{\mathrm{EM}}\left(\hat{y}, y^{*}\right)=\left\{\begin{array}{ll}1, & \text { if } \hat{y}=y^{*} \\0, & \text { otherwise }\end{array}\right.
\end{equation}

This strict binary reward provides an unambiguous reward signal, encouraging the agenic LLM to produce fully correct answers rather than partially correct or ambiguous outputs. Compared to continuous value rewards like F1-score, it has a higher upper bound for reinforcement training.

\textbf{Format Constraint} In addition to the correctness of answer, we enforce a format validation constraint on the model output as well. Specifically, the generated response is necessary to contain both a reasoning trace and a final answer segment, explicitly delimited by the tags \textit{<think>...</think>} and \textit{<answer>...</answer>}, respectively. Besides, The final answer used for matching needs to be enclosed within \textit{\textbackslash{}boxed\{\}} (within \textit{<answer>...</answer>}). Formally, we define a format validation function as follows:
\begin{equation}
\mathbb{I}_{\text {format }}=\left\{\begin{array}{ll}1, & \text { if both tags are present } \\0, & \text { otherwise }\end{array}\right.
\end{equation}

If the response violates the required format, the reward is set to zero regardless of the predicted answer correctness. This plays an important role for improving the model's ability to follow user's instructions.

\textbf{Final Reward Definition} Based on the above, the overall reward $r$ is defined as the synthesis of exact match correctness and format constraint:
\begin{equation}
r=\left\{\begin{array}{ll}r_{\mathrm{EM}}(\hat{y},y^*), & \text { if $\mathbb{I}_{\text {format }}=1$ } \\-1, & \text { otherwise }\end{array}\right.
\end{equation}
Consequently, the agentic LLM derives a reward of 1 if and only if it outputs a correctly formatted output whose final answer exactly matches the ground truth.

\begin{figure*}[t]
\centering
\begin{tcolorbox}[
    colback=white,
    colframe=gray,
    arc=10pt,
    boxrule=3pt,
    left=10pt,
    right=10pt,
    top=5pt,
    bottom=5pt,
    toptitle=3pt,
    bottomtitle=3pt,
    title={\textcolor{white}{\normalsize\bfseries Prompt Template.}},
    colbacktitle=headercolor,
    coltitle=white,
    fonttitle=\bfseries\Large
]

You are a helpful assistant that can solve the given question step by step with the help of the wikipedia search tool. Given a question, you need to first think about the reasoning process in the mind and then provide the answer. During thinking, you can invoke the wikipedia search tool to search for fact information about specific topics if needed. You can search as many times as your want. The reasoning process and answer are enclosed within \textcolor{myblue}{<think> </think>} and \textcolor{mypurple}{<answer> </answer>} tags respectively, and the search query and result are enclosed within \textcolor{myred}{<search> </search>} and \textcolor{mygreen}{<result> </result>} tags respectively. For example, \textcolor{myblue}{<think>} This is the reasoning process. \textcolor{myblue}{</think>} \textcolor{myred}{<search>} search query here \textcolor{myred}{</search>} \textcolor{mygreen}{<result>} search result here \textcolor{mygreen}{</result>} \textcolor{myblue}{<think>} This is the reasoning process. \textcolor{myblue}{</think>} \textcolor{mypurple}{<answer>} The final answer is [ \textbf{\textbackslash{}boxed\{answer here\}} ] \textcolor{mypurple}{</answer>}. In the last part of the answer, the final exact answer is enclosed within \textbf{\textbackslash{}boxed\{\}} with latex format. Question:

\end{tcolorbox}
\caption{The prompt template.}
\label{prompt template}
\end{figure*}

\subsection{Prompt Template}

We present our prompt template as Figure~\ref{prompt template}. We follow the similar setting in \cite{jin2025search} and \cite{dong2025agenticaepo} with different tags containing corresponding content. Specifically, the model outputs its intermediate reasoning within the \textcolor{myblue}{\textit{<think></think>}} tags and makes a search request by emitting the \textcolor{myred}{\textit{<search></search>}} tag, which corresponds to an action in the agent–environment interaction process. The content returned by the search engine is wrapped in \textcolor{mygreen}{\textit{<result></result>}} tags and embedded into the response context. Finally, after completing the reasoning process, the model outputs the content of answer within the \textcolor{mypurple}{\textit{<answer></answer>}} tags. Noting that we exact the final answer from the \textbf{\textbackslash{}boxed\{\}} in response to achieve more accurate EM score matching.

\subsection{Datasets}
We introduce two categories of benchmarks adopted in our experiments in this section.

\textbf{Multi-Hop QA}. To evaluate the multi-turn tool calling and complex reasoning capabilities, we adopt four datasets that require reasoning over multiple steps. \textbf{HotpotQA} \cite{yang2018hotpotqa} is a large-scale Wikipedia-based benchmark with strong supervision for supporting facts, widely used to evaluate explainable multi-hop reasoning. \textbf{2WikiMultiHopQA} \cite{ho2020constructing} consists of Wikipedia text and Wikidata triples, which is a comprehensive benchmark with multi-hop dependency between entities. \textbf{Musique} \cite{trivedi2022musique} has 25k 2-4 hop questions synthesized via controlled composition of single-hop content. \textbf{Bamboogle} \cite{press2023measuring} is a tiny dataset consisting of complex compositional QA questions, which benefits the evaluation of the effectiveness and stability of agentic RL methods.

\noindent \textbf{Single-Hop QA}. We further adopt three single-hop QA benchmarks. \textbf{Natural Questions (NQ)} \cite{kwiatkowski2019natural} is widely used to assess retrieval-augmented generation, which includes large quantity of user queries from Wikipedia. \textbf{TriviaQA} \cite{joshi2017triviaqa} contains questions with syntactic and lexical variability between questions and corresponding answer-evidence sentences. \textbf{PopQA} \cite{mallen2023llm_memorization} is an entity-centric open-domain QA dataset, designed to explore the interaction between agentic retrieval and parametric memorization.

\subsection{Baseline Settings}
In Table~\ref{tab:hyperparameters}, we exhitbit common hyperparameter settings used for baselines in main experiment. 
We further detail the baselines selected in our main experiments and illustrate the corresponding specially adjusted experiment settings. 

\begin{itemize}
    \item \textbf{ReAct} \cite{yao2023react}: A prompting paradigm to enable procedural planning and dynamic function calls(tool calls), which is training-free.
    \item \textbf{GRPO} \cite{guo2025deepseek}: A group-based on-policy RL optimization method, which generates n sets of candidates by multiple sampling for each prompt to derive the relative advantages, thereby replacing the traditional critic used in PPO. We set the coefficient to 0.001 for KL constraint. The clip ratio is set to 0.2.
    \item \textbf{DAPO}~\cite{yu2025dapo}: An extension of GRPO that incorporates a decoupled clipping mechanism and dynamic sampling to stabilize policy updates. Following the recommended configuration, we set \texttt{clip\_ratio\_low} and \texttt{clip\_ratio\_high} to \(0.2\) and \(0.28\), respectively, to enable a wider clipping range. Additionally, we configure the \texttt{overlong\_buffer} with a capacity of \(2000\) (penalty factor \(1.0\)) and activate the dynamic sampling mechanism. Consistent with DAPO’s default setup, the generation batch size is set to three times the training batch size. 
    \item \textbf{GSPO}~\cite{zheng2025groupsequencepolicyoptimization}: A sequence-level variant of GRPO that defines importance ratios and clipping based on full response likelihoods, trading fine-grained token-level control for improved sequence-level stability. Analogous to DAPO, we adopt asymmetric clipping with \texttt{clip\_ratio\_low} = \(3\!\times\!10^{-4}\) and \texttt{clip\_ratio\_high} = \(4\!\times\!10^{-4}\) to support precise sequence-level clipping.
    \item \textbf{AEPO}~\cite{dong2025agenticaepo}: A state-of-the-art agentic RL method that introduces entropy-balanced rollout scheduling and entropy-aware clipping to mitigate over-branching and gradient collapse in tree-based rollouts. In our experiments, we set \texttt{initial\_rollouts} to \(8\). Additional hyperparameters include \texttt{beam\_size} = \(2\), \texttt{branch\_probability} = \(0.5\), and \texttt{entropy\_weight} = \(0.2\).
    \item \textbf{Tree-GRPO}~\cite{ji2025treesearchllmagent}: A tree-based agentic RL framework that integrates GRPO with tree search to enable fine-grained credit assignment across branching trajectories. We replicate the original experiment configuration as reported in the paper.
\end{itemize}

\begin{table}[t]
\centering
\small
\setlength\tabcolsep{8pt}
\renewcommand\arraystretch{1.2}
\begin{tabular}{lcccccccc}
\hline
Config & Value  \\
\hline
optimizer & AdamW  \\
learning rate & 1e-6  \\
clip\_ratio & 0.2 \\
total training steps & 240 \\
training batch size & 64 \\
PPO mini batch size & 8 \\
rollout\_n & 16 \\
max prompt length & 2000 \\
max response length & 6192  \\
max tool-calls & 6 \\
reward metrics & EM  \\
retriever & local wiki \\
top-K retrieval passages & 3 \\
\hline
\end{tabular}
\caption{General hyperparameters of baseline methods in main experiments.}
\label{tab:hyperparameters}
\end{table}

\subsection{Search Tool Settings}
Follow the settings of search tool in Search-R1~\cite{jin2025search}, we utilize the wikipedia as the corpus for search engine with e5-base-v2~\cite{wang2022text} as the retriever model. The retrieval knowledge base contains 21 million data entries obtained from Wikipedia. After the agentic LLM calls the wiki search tool, the search engine will return the top-k most relevant knowledge data determined by the retrieval model at each turn.

\subsection{Hardware and Artifacts}
All experiments are conducted on a system equipped with 8 NVIDIA H20 GPUs. As the backbone models, we employ Qwen3-4B, Qwen3-8B, and Qwen2.5-7B~\cite{yang2025qwen3technicalreport, qwen2025qwen25technicalreport}, chosen for their strong foundational reasoning capabilities in agentic RL training. Our implementation is built on the VeRL framework~\cite{sheng2024hybridflow, zhang2024framework}, a mature and stable RL infrastructure that can be easily integrated with agentic RL training features.

\section{Analysis on Training Stability}

\label{Sec:infra}
\begin{figure}[h]
    \centering
    \includegraphics[width=0.49\textwidth]{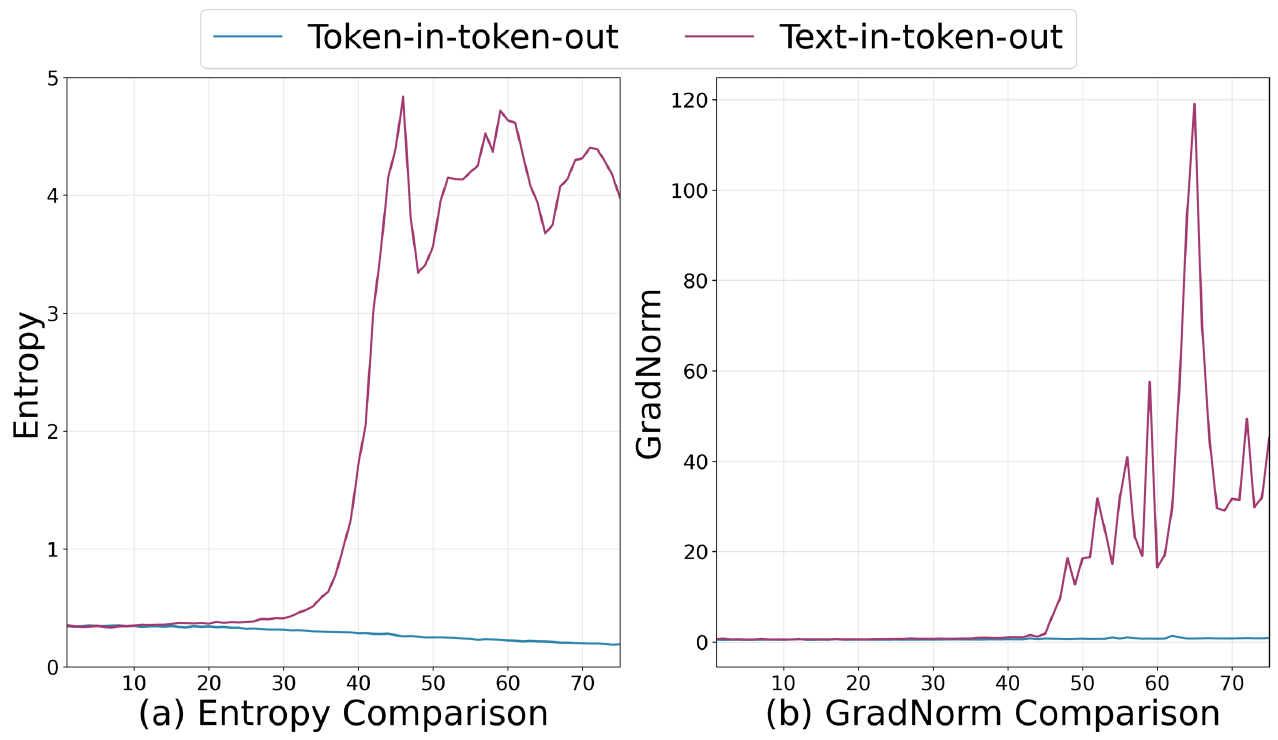}
    \caption{Training comparison between the scheme of token-in-token-out and text-in-token-out on Qwen3-4B with multi-hop training set.}
    \label{fig:infra-combined}
\end{figure}

\begin{figure*}[t]
\captionsetup{type=table}  \small
    \centering
    \begin{tabular}{ccccccc}
        \toprule
        \textbf{Advantage $A(n)$} & \textbf{Value $V(n)$}&\textbf{Hotpot}&\textbf{2wiki}&\textbf{Musiq}&\textbf{Bamb}&\textbf{Avg.} \\
        \midrule
         $V(n)$ & Child Weighted &\textbf{49.44}&52.99&\textbf{24.80}&\textbf{54.4}& \textbf{48.81} \\
         $V(n)$ & Child Mean &47.95&\textbf{53.45}&23.56&52.8& 48.43 \\
         $V(n)$ & Leaf Mean &49.10&52.76&24.18&54.4& 48.52 \\
         \midrule
         $L(n)=V(n)-V(n_{\text{parent}})$ & Child Weighted &44.73&51.38&19.73&51.2& 45.80 \\
         $L(n)=V(n)-V(n_{\text{parent}})$ & Leaf Mean &45.36&50.53&20.40&52.8&45.61  \\
         \midrule
         $G(n)=V(n)-V(n_{\text{root}})$ & Child Weighted &48.45&52.87&22.73&54.4& 48.19 \\
         $G(n)=V(n)-V(n_{\text{root}})$ & Leaf Mean &47.67&50.96&23.68&52.8&46.98  \\
         \midrule
         $L(n)+G(n)$ & Leaf Mean &45.35&48.24&19.07&52.0&44.18  \\
        \bottomrule
    \end{tabular}

    \caption{Ablation on the effectiveness of different turn-wise credit assignment schemes.}
    \label{tab:stepwise2}

\captionsetup{type=figure}
    \centering
    \includegraphics[width=0.95\textwidth]{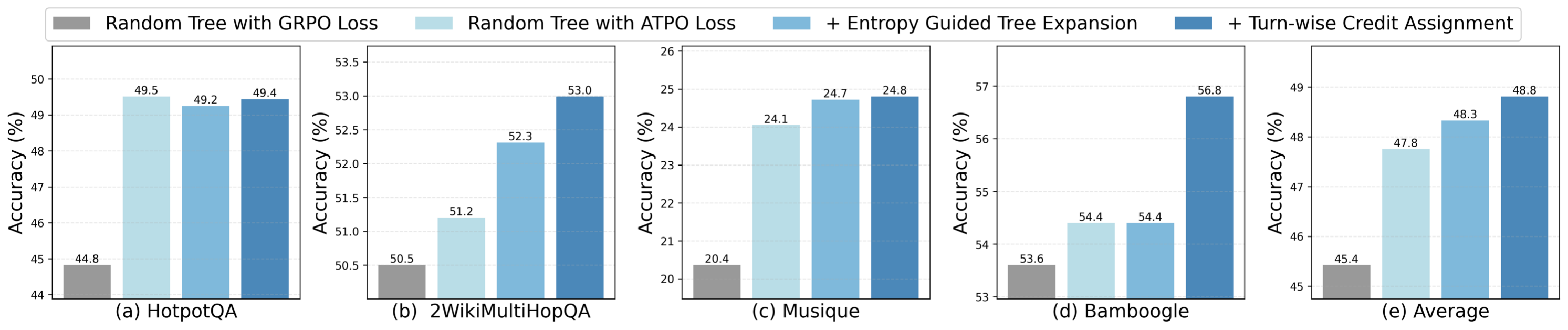}
    \caption{Ablation Study on progressive molular designs applied to AT\textsuperscript{2}PO.}
    \label{fig:ablation}
\end{figure*}

During our replication of Tree-GRPO~\cite{ji2025treesearchllmagent} using the open-sourced implementation, we observed significant training instability across multiple runs, often leading to complete collapse at very early stages when using Qwen3-based models. We trace this issue primarily to a phenomenon known as Retokenization Drift~\cite{vllm2025retokenization}.

In agentic RL rollouts in the Tree-GRPO implementation, intermediate responses are first detokenized into text strings to enable tool invocation and parsing. Tool outputs in string form are then concatenated to the dialogue history, and the entire sequence is retokenized before being used for policy updates. Despite semantic equivalence at the string level, this text-in–token-out pipeline can yield different token sequences before and after retokenization due to tokenizer nondeterminism, introducing critical inconsistencies between the sampled and training trajectories.

To address this, we adopt a token-in–token-out approach: we preserve the exact token IDs generated at each turn during rollout and use them directly in training, bypassing intermediate detokenization and retokenization. As illustrated in Figure~\ref{fig:infra-combined}, this modification eliminates retokenization-induced distributional shifts. Empirically, the revised pipeline stabilizes training—both policy entropy and gradient norm remain consistent throughout training, with no signs of collapse.

While this fix resolves a key source of instability, other challenges in stabilizing agentic RL training persist, including consistency between training and inference, etc. These remain important directions for future work beyond the scope of this paper.



\section{Detailed Ablations}
\subsection{Ablation Study on AT\textsuperscript{2}PO}

Table~\ref{tab:ablation} and Figure~\ref{fig:ablation} present an ablation study on Qwen3-4B to assess the contribution of each component in AT\textsuperscript{2}PO, following a progressive modular where modules are incrementally added. This setup enables us to isolate the effect of each design choice under a fixed backbone and consistent task suite.

Starting from a base configuration that uses the standard GRPO loss with random tree search, replacing it with our proposed ATPO loss yields substantial gains across all multi-hop benchmarks. This highlights that the turn-wise importance sampling and clipping mechanism provides a significantly finer-grained optimization objective than vanilla GRPO in multi-turn agentic RL. Next, integrating Entropy-Guided Tree Expansion leads to a further clear performance improvement. This indicates that entropy-aware branching facilitates more effective exploration during tree-based rollouts, enabling the agent to allocate its limited search budget toward more informative and uncertain decision points. Finally, adding Turn-wise Credit Assignment—completing the full AT\textsuperscript{2}PO framework—delivers additional gains and achieves the best overall performance. This demonstrates that providing fine-grained, step-level learning signals to individual turns effectively mitigates the credit assignment problem inherent in sparse-reward, long-horizon agentic tasks.

\subsection{Scheme Analysis of Turn-wise Credit Assignment}

Table~\ref{tab:stepwise} and Table~\ref{tab:stepwise2} compare different formulations of turn-wise credit assignment by varying both the advantage definition and the value aggregation strategy.
We consider three strategies for estimating the node value $V_n$: (a) child-weighted propagation as defined in in Eq~(3), (b) mean of all child values and (c) mean of all leaf values following TreeRL~\cite{hou2025treerl}. As for advantage computation $f(\cdot)$, our main experiments adopt the simplest formulation $A_n=V_n$. We further compare this with several difference-based alternatives: (a) the TD advantage $A_n=L_n=V_n-V_{n_{\text{parent}}}$, (b) the global advantage $A_n=G_n=V_n-V_{n_{\text{root}}}$ and (c) the combined signal $A_n=L_n+G_n$ as used in prior work~\cite{hou2025treerl}. 

Overall, directly setting the advantage equal to the node value yields consistently strong performance across all datasets. Under this formulation, child-weighted aggregation achieves the best results, while child-mean and leaf-mean aggregation remain competitive but slightly inferior. In contrast, difference-based formulations lead to noticeably lower performance. Surprisingly, the combined signal, despite its demonstrated effectiveness in conventional RLVR settings~\cite{yao2023tree}, performs worst among all variants. In the agentic RL scenario, directly using $V_n$ as the advantage provides a cleaner and more compatible learning signal.

\section{Case Study}
We conduct the case study which is presented through Table~\ref{casestudy1} to Table~\ref{casestudy7}. Specifically, in order to more intuitively illustrate the reasoning process and tool usage in agentic RL, we selected one representative case generated by our method from each of the seven benchmarks used in the main experiments above. We use different colored boxes to describe the different states and actions of the agentic LLM during the reasoning process. For multi-hop QA datasets, which mainly include more complex question-answering problems requiring multi-step reasoning, agentic LLMs typically exhibit more multi-turn tool calls and interactions compared to single-hop QA questions. This is consistent with the statistics shown in Figure~\ref{fig:turn_dist}.

\begin{table*}[!t]
    \centering
    \fontsize{9pt}{11pt}\selectfont
    \begin{tabular}{p{0.98\linewidth}}
    \midrule
        \rowcolor{gray!20}\textbf{Example \#1 from HotpotQA Dataset} \\
    \midrule
        \textbf{Question:} \\
        What was the father of Kasper Schmeichel voted to be by the IFFHS in 1992?\\
        \textbf{Labeled Answer:} \\
        World's Best Goalkeeper \\
    \midrule
        \rowcolor{gray!20}
        \textbf{Model Output by AT\textsuperscript{2}PO(Qwen3-8B)}\\
    \midrule
        \blue{<think>} Okay, let's try to figure out what the father of Kasper Schmeichel was voted to be by the IFFHS in 1992. First, I need to identify who Kasper Schmeichel's father is. I know that Kasper Schmeichel is a Danish footballer, a goalkeeper. So, maybe his father is also a footballer? I should start by searching for Kasper Schmeichel's father. Let me use the Wikipedia search tool for that. Let me search "Kasper Schmeichel father".\blue{</think>}
        \\\red{<search>} Kasper Schmeichel father \red{</search>}
        \\\green{<result>}\\
Page 1: "Kasper Schmeichel" surpassing his father's record. During the Round of 16 match against Croatia, Schmeichel was named Man of the Match after saving a penalty-kick taken by Luka Modrić in extra-time and two penalties during the subsequent shootout, although his opposing counterpart Danijel Subašić saved three penalties in that shootout to eliminate Denmark. Schmeichel is the son of former Denmark international goalkeeper Peter Schmeichel. Having spent most of his childhood living in England due to his father's playing career, Kasper is a native speaker of both Danish and English, the latter of which he speaks with a Northern-English accent.  \\
Page 2: "Peter Schmeichel" "125 greatest living footballers", at the FIFA 100 celebrations. His son, Kasper, is also a professional football goalkeeper, currently playing for Premier League side Leicester City and the Danish national team. Peter Schmeichel was born in the Søborggård parish of Gladsaxe, Denmark, to Inger, a Danish nurse, and Antoni Schmeichel, a Polish jazz musician. He held Polish citizenship until 1970 when he, his father, and his three sisters became Danish citizens. Schmeichel inherited his middle name – Bolesław – from his great-grandfather. He spent his early years in the town of Buddinge, Copenhagen.  \\
Page 3: "Kasper Schmeichel" Hall Grammar School in Cheadle Hulme, Stockport, Greater Manchester, and he also attended Saint Julian's School in Portugal while his father played for Sporting CP in Lisbon. When Alex Bruce, son of Kasper's father's former Manchester United teammate Steve Bruce, joined Kasper at Leeds United, the move saw the two players re-unite after both had previously played football together on the streets as children after school. Schmeichel has two children: a son, Max (born 2010) and a daughter, Isabella (born 2012). Schmeichel signed an ambassadorial deal with Precision Goalkeeping on 17 August 2010.
        \\\green{</result>} 
        \\\blue{<think>} Okay, so the search result says that Kasper Schmeichel's father is Peter Schmeichel. Now, the question is about what Peter Schmeichel was voted to be by the IFFHS in 1992. I need to check the IFFHS awards or something related to Peter Schmeichel in 1992. First, I should search who Peter Schmeichel is. Wait, I already have that from the result. Let me search for IFFHS and Peter Schmeichel in 1992. \blue{</think>}
        \\\red{<search>} Peter Schmeichel IFFHS 1992 \red{</search>}
        \\\green{<result>}\\
        \textbf{***One turn search results omitted here***}
        \\\green{<result>}\\
        \blue{<think>} Okay, so the search result says that Peter Schmeichel was voted the IFFHS World's Best Goalkeeper in 1992. The father of Kasper Schmeichel is Peter Schmeichel, a renowned Danish former professional footballer who played as a goalkeeper. According to the information retrieved, Peter Schmeichel was voted the IFFHS World's Best Goalkeeper in 1992. So the answer is that his father was voted as the IFFHS World's Best Goalkeeper in 1992. \blue{</think>}
        \\\purple{<answer>} \\ 
        \boxed{\text{World's Best Goalkeeper}}
        \\\purple{</answer>}\\
    \bottomrule
    \caption{An example from AT\textsuperscript{2}PO on HotpotQA dataset, with special symbols used in think content, search queries, returned results and final answer highlighted with \blue{blue box}, \red{red box}, \green{green box} and \purple{purple box}, respectively.}
    \label{casestudy1}
    \end{tabular}
\end{table*}

\begin{table*}[!t]
    \centering
    \fontsize{9pt}{11pt}\selectfont
    \begin{tabular}{p{0.98\linewidth}}
    \midrule
        \rowcolor{gray!20}\textbf{Example \#1 from Musique Dataset} \\
    \midrule
        \textbf{Question:} \\
        Which company owns the manufacturer of Learjet 60?\\
        \textbf{Labeled Answer:} \\
        Bombardier Inc/Bombardier \\
    \midrule
        \rowcolor{gray!20}
        \textbf{Model Output by AT\textsuperscript{2}PO(Qwen3-8B)}\\
    \midrule
        \blue{<think>} Okay, so I need to find out which company owns the manufacturer of the Learjet 60. Let me start by recalling what I know. Learjet is a brand of business jets, right? I think Learjet was originally a company that made these jets. But I'm not sure about the current ownership. First, I should figure out who manufactures the Learjet 60. Maybe I should search for the manufacturer of the Learjet 60. Let me do that. I'll use the Wikipedia search tool to look up the Learjet 60.\blue{</think>}
        \\\red{<search>} Learjet 60 manufacturer \red{</search>}
        \\\green{<result>}\\
Page 1: "Learjet 60" Learjet 60 The Learjet 60 is a mid-size cabin, medium-range business jet aircraft manufactured by Bombardier Aerospace in Wichita, Kansas. Powered by two Pratt \& Whitney Canada PW305A engines, it has a range (with 4 passengers and 2 crew) of with NBAA reserves, ISA. In July 2012 Bombardier Aerospace announced a temporary "production pause" of the latest variant Learjet 60XR to begin in the fourth quarter of 2012. The Learjet 60 is an improved version of the Learjet 55, with a longer fuselage and more powerful turbofan engines. It first flew on 10 October 1990 and received FAA certification.   \\
Page 2: "Learjet 60" and fractional jet operators. Learjet 60 The Learjet 60 is a mid-size cabin, medium-range business jet aircraft manufactured by Bombardier Aerospace in Wichita, Kansas. Powered by two Pratt \& Whitney Canada PW305A engines, it has a range (with 4 passengers and 2 crew) of with NBAA reserves, ISA. In July 2012 Bombardier Aerospace announced a temporary "production pause" of the latest variant Learjet 60XR to begin in the fourth quarter of 2012. The Learjet 60 is an improved version of the Learjet 55, with a longer fuselage and more powerful turbofan engines. It first flew on 10 October 1990.   \\
Page 3: "Learjet 60" It also distinguished as the last legacy Learjet, using a variation of the wing that designer Bill Lear adapted from the Swiss military aircraft, the FFA P-16. The next-generation Learjet was to be the Learjet 85 and was an all-new design by Bombardier Aerospace slated for delivery in 2013. The Learjet 60, while a tremendous performer, also maintains the highest incident/accident rate in its class with most accidents occurring during landing. The use of the original Lear 23 gear and wheels (albeit it with an added brake rotor, bringing the total to 3 per wheel assembly).
        \\\green{</result>} 
        \\\blue{<think>} Okay, so the Learjet 60 is manufactured by Bombardier Aerospace. Now the question is asking which company owns the manufacturer. So Bombardier Aerospace is the manufacturer. Therefore, I need to find out who owns Bombardier Aerospace. Let me search for Bombardier Aerospace's parent company. \blue{</think>}
        \\\red{<search>} Bombardier Aerospace parent company \red{</search>}
        \\\green{<result>}\\
        \textbf{***One turn search results omitted here***}
        \\\green{<result>}\\
        \blue{<think>} Okay, according to this, Bombardier Aerospace is a division of Bombardier Inc. So the manufacturer of the Learjet 60 is Bombardier Aerospace, which is owned by Bombardier Inc. Therefore, the answer should be Bombardier Inc. \blue{</think>}
        \\\purple{<answer>} \\ 
        \boxed{\text{Bombardier Inc}}
        \\\purple{</answer>}\\
    \bottomrule
    \caption{An example from AT\textsuperscript{2}PO on Musique dataset, with special symbols used in think content, search queries, returned results and final answer highlighted with \blue{blue box}, \red{red box}, \green{green box} and \purple{purple box}, respectively.}
    \label{casestudy2}
    \end{tabular}
\end{table*}

\begin{table*}[!t]
    \centering
    \fontsize{9pt}{11pt}\selectfont
    \begin{tabular}{p{0.98\linewidth}}
    \midrule
        \rowcolor{gray!20}\textbf{Example \#1 from 2WikiMultiHopQA Dataset} \\
    \midrule
        \textbf{Question:} \\
        Which film was released earlier, August Underground'S Penance or Vasantha Raagam?\\
        \textbf{Labeled Answer:} \\
        Vasantha Raagam \\
    \midrule
        \rowcolor{gray!20}
        \textbf{Model Output by AT\textsuperscript{2}PO(Qwen3-8B)}\\
    \midrule
        \blue{<think>} Okay, I need to find out which film was released earlier between "August Underground's Penance" and "Vasantha Raagam." Let me start by recalling what I know about these two films. First, "Vasantha Raagam" might be a South Indian film, possibly from Tamil or Telugu cinema, given the name. I remember that "Vasantha Raagam" is a classic film, maybe from the 1960s or 1970s. But I'm not entirely sure. On the other hand, "August Underground's Penance" sounds like it could be an independent or arthouse film. The title seems a bit more modern. However, I'm not certain about its release date. To get accurate information, I should use the Wikipedia search tool to find the release dates of both films. Let me start by searching for "Vasantha Raagam" to find its release year.\blue{</think>}
        \\\red{<search>} Vasantha Raagam \red{</search>}
        \\\green{<result>}\\
Page 1: "Vasantam" Vasantam is a Tollywood film produced by N.V.Prasad \& Sanam Naga Ashok kumar on Sri Sai Deva Productions banner, directed by Vikraman. Starring Venkatesh, Arti Agarwal, Kalyani in lead roles and music is composed by S. A. Rajkumar.The film won for two Nandi Awards. The film opened simultaneously alongside the Tamil version of the film, "Priyamaana Thozhi". Ashok (Venkatesh) and Julie (Kalyani) are childhood friends. Ashok is like a family member of Julie's family. They are very close and believe that friendship is above love, yet they do not have any love feelings towards each other.       \\
Page 2: "Vasantham Vanthachu" Vasantham Vanthachu is a 2007 Tamil romantic comedy film directed by K. Viveka Bharathi. The film features Venkat Prabhu and Nanditha Jennifer in lead roles, with Shanmugasundaram, Vaiyapuri, Ramji, Varalakshmi, Sabitha Anand, Vasu Vikram and Velu Thigal playing supporting roles. The film, produced by Velu Thigal and Amba Umadhevi, had musical score by Kavin Saradha and Raj Shankar, and was released on 14 September 2007. Kumaresan (Venkat Prabhu) is a milk vendor in his village, who hates the marriage and stays away from women. His elderly father (Velu Thigal) wants him to get married as soon as possible.  \\
Page 3: "Vasanth" Vasanth is an Indian film director and screenwriter, working in the Tamil film industry. Following a stint as journalist and after assisting K. Balachander, he made his directorial debut with "Keladi Kanmani" in 1990, which along with the trendsetting "Aasai" (1995), are considered his finest and most successful films. Known for being realistic, his films have been described as a "middle path" between parallel and commercial cinema, revolving around social themes and relationships. Vasanth began his career as a journalist and short story writer, before he began working as an assistant director to K. Balachander on 18 films.
        \\\green{</result>} 
        \\\textbf{***Four turn search steps omitted here***}
        \\\blue{<think>} The search result doesn't mention "Vasantha Raagam" directly. However, given that "August Underground's Penance" is from 2007, and "Vasantha Raagam" might be a 1960s film, I can infer that "Vasantha Raagam" was released earlier. Alternatively, if "Vasantha Raagam" is a 2007 film, then both would be the same. But since the user is asking which was released earlier, and "August Underground's Penance" is 2007, I need to check if "Vasantha Raagam" is older. Since I couldn't find exact info, I'll have to make an educated guess based on the available data. Given that "August Underground's Penance" is from 2007, and "Vasantha Raagam" might be a classic film from the 1960s or 1970s, it's likely that "Vasantha Raagam" was released earlier. Based on the information gathered, "August Underground's Penance" was released in 2007. "Vasantha Raagam" is likely a classic film from the 1960s or 1970s, which would make it older than "August Underground's Penance". \\\blue{</think>}
        \\\purple{<answer>} \\ 
        \boxed{\text{Vasantha Raagam}}
        \\\purple{</answer>}\\
    \bottomrule
    \caption{An example from AT\textsuperscript{2}PO on 2WikiMultiHopQA dataset, with special symbols used in think content, search queries, returned results and final answer highlighted with \blue{blue box}, \red{red box}, \green{green box} and \purple{purple box}, respectively.}
    \label{casestudy3}
    \end{tabular}
\end{table*}

\begin{table*}[!t]
    \centering
    \fontsize{9pt}{11pt}\selectfont
    \begin{tabular}{p{0.98\linewidth}}
    \midrule
        \rowcolor{gray!20}\textbf{Example \#1 from Bamboogle Dataset} \\
    \midrule
        \textbf{Question:} \\
        Who was the first African American mayor of the most populous city in the United States?\\
        \textbf{Labeled Answer:} \\
        David Dinkins \\
    \midrule
        \rowcolor{gray!20}
        \textbf{Model Output by AT\textsuperscript{2}PO(Qwen3-8B)}\\
    \midrule
        \blue{<think>} Okay, so I need to find out who was the first African American mayor of the most populous city in the United States. Let me break this down step by step. First, the most populous city in the United States. I remember that this is usually a city like New York City, but I should confirm that. Let me do a quick search to make sure.\blue{</think>}
        \\\red{<search>} most populous city in the United States \red{</search>}
        \\\green{<result>}\\
Page 1: "New York City" New York City, often called New York City (NYC) or simply New York (NY), is the most populous city in the United States. With an estimated 2017 population of 8,622,698 distributed over a land area of about, New York City is also the most densely populated major city in the United States. Located at the southern tip of the state of New York, the city is the center of the New York metropolitan area, the largest metropolitan area in the world by urban landmass and one of the world's most populous megacities.     \\
Page 2: "Demography of the United States" its southern coast - with particular emphasis on the city of Anchorage - and Hawaii's is centered on the island of Oahu. California and Texas are the most populous states, as the mean center of U.S. population has consistently shifted westward and southward. New York City is the most populous city in the United States. The United States Census Bureau shows a population increase of 0.75\% for the twelve-month period ending in July 2012. Though high by industrialized country standards, this is below the world average annual rate of 1.1\%.  \\
Page 3: \textbf{***more search results  omitted here***}
        \\\green{</result>} 
        \\\textbf{***Five turn search steps omitted here***}
        \\\blue{<think>} Okay, so the most populous city is New York City. Now, the next part is to find the first African American mayor of that city. Let me search for that. \blue{</think>}
        \\\red{<search>} first African American mayor of New York City \red{</search>}
        \\\green{<result>}
        \\\textbf{***One turn search results omitted here***}
        \\\green{</result>}
        \\\textbf{***Five turn search steps omitted here***}
        \\\blue{<think>}
        Okay, the search result says that David Dinkins was the first and only African American mayor of New York City. The most populous city in the United States is New York City. After searching for the first African American mayor of New York City, the information consistently points to David Dinkins as the first and only African American mayor of the city, serving from 1990 to 1993. So the answer should be David Dinkins.
        \\\blue{</think>}
        \\\purple{<answer>} \\ 
        \boxed{\text{David Dinkins}}
        \\\purple{</answer>}\\
    \bottomrule
    \caption{An example from AT\textsuperscript{2}PO on Bamboogle dataset, with special symbols used in think content, search queries, returned results and final answer highlighted with \blue{blue box}, \red{red box}, \green{green box} and \purple{purple box}, respectively.}
    \label{casestudy4}
    \end{tabular}
\end{table*}

\begin{table*}[!t]
    \centering
    \fontsize{9pt}{11pt}\selectfont
    \begin{tabular}{p{0.98\linewidth}}
    \midrule
        \rowcolor{gray!20}\textbf{Example \#1 from NQ Dataset} \\
    \midrule
        \textbf{Question:} \\
        how many episodes are there in dragon ball z?\\
        \textbf{Labeled Answer:} \\
        291 episodes/291 \\
    \midrule
        \rowcolor{gray!20}
        \textbf{Model Output by AT\textsuperscript{2}PO(Qwen3-8B)}\\
    \midrule
        \blue{<think>} Okay, so the user is asking how many episodes there are in Dragon Ball Z. Hmm, I need to figure this out. Let me start by recalling what I know. Dragon Ball Z is a popular anime series that's part of the Dragon Ball franchise. It's known for having a lot of episodes, maybe over 200? But I need to confirm that. Let me use the Wikipedia search tool to get accurate data. First, I'll search for "how many episodes are there in dragon ball z" using the tool. \blue{</think>}
        \\\red{<search>} how many episodes are there in dragon ball z \red{</search>}
        \\\green{<result>}\\
Page 1: "Dragon Ball Z" It premiered in Japan on Fuji Television on April 26, 1989, taking over its predecessor's time slot, and ran for 291 episodes until its conclusion on January 31, 1996. Because Toriyama was writing the manga during the production of the anime, "Dragon Ball Z" added original material not adapted from the manga, including lengthening scenes or adding new ones, and adding new attacks and characters not present in the manga. For example, Toriyama was asked to create an additional character for Goku's training with King Kai, resulting in the cricket Gregory.    \\
Page 2: "Dragon Ball" April 5, 2009, the series premiered in Japan airing in Fuji TV. "Dragon Ball Z Kai" reduced the episode count to 159 episodes (167 episodes internationally), from the original footage of 291 episodes. Damaged frames were removed, resulting in some minor shots being remade from scratch in order to fix cropping, and others to address continuity issues. The majority of the international versions, including Funimation Entertainment's English dub, are titled "Dragon Ball Z Kai". premiered on Fuji TV on February 7, 1996 and ran until November 19, 1997 for 64 episodes.  \\
Page 3: "Dragon Ball" both based on the same outline by Toriyama for the story of "Super". Toei Animation produced an anime television series based on the first 194 manga chapters, also titled "Dragon Ball". The series premiered in Japan on Fuji Television on February 26, 1986 and ran until April 19, 1989, lasting 153 episodes. Instead of continuing the anime as "Dragon Ball", Toei Animation decided to carry on with their adaptation under a new name and asked Akira Toriyama to come up with the title.
        \\\green{</result>} 
        \\\blue{<think>} Okay, the first result says that Dragon Ball Z ran for 291 episodes from April 26, 1989, to January 31, 1996. That seems like the main series. But the second result mentions "Dragon Ball Z Kai" which reduced the episode count to 159 episodes (167 internationally). However, the user is asking about the original Dragon Ball Z, not the Kai version. The third result talks about the original Dragon Ball series, which had 153 episodes, but that's before Z. So the answer is 291 episodes. \\\blue{</think>}
        \\\purple{<answer>} \\ 
        \boxed{\text{291 episodes}}
        \\\purple{</answer>}\\
    \bottomrule
    \caption{An example from AT\textsuperscript{2}PO on NQ dataset, with special symbols used in think content, search queries, returned results and final answer highlighted with \blue{blue box}, \red{red box}, \green{green box} and \purple{purple box}, respectively.}
    \label{casestudy5}
    \end{tabular}
\end{table*}

\begin{table*}[!t]
    \centering
    \fontsize{9pt}{11pt}\selectfont
    \begin{tabular}{p{0.98\linewidth}}
    \midrule
        \rowcolor{gray!20}\textbf{Example \#1 from TriviaQA Dataset} \\
    \midrule
        \textbf{Question:} \\
        Who had an 80s No 1 hit with Hold On To The Nights?\\
        \textbf{Labeled Answer:} \\
        Richard Noel Marx/Richard Marx \\
    \midrule
        \rowcolor{gray!20}
        \textbf{Model Output by AT\textsuperscript{2}PO(Qwen3-8B)}\\
    \midrule
        \blue{<think>} Okay, so the question is asking who had an 80s No. 1 hit with "Hold On To The Nights." Hmm, I need to figure this out. Let me start by recalling if I know the answer. I think it might be a band from the 80s, maybe a British group? Or maybe a singer? Let me check using the Wikipedia search tool to confirm. First, I'll search for "Hold On To The Nights 80s No 1 hit" to see if that gives me any clues. \blue{</think>}
        \\\red{<search>} who had an 80s no 1 hit with hold on to the nights \red{</search>}
        \\\green{<result>}\\
Page 1: "Hold On to the Nights" Hold On to the Nights "Hold On to the Nights" is a power ballad by American rock singer/songwriter/musician Richard Marx. This was the fourth and final single released from his self-titled debut album, and the first to reach number-one on the US "Billboard" Hot 100 chart. The song has been re-released on numerous albums and is included on Marx's live performance DVD "A Night Out with Friends" (2012). Written by Richard Marx, "Hold On to the Nights" reached the "Billboard" Hot 100 number 1 position on July 23, 1988, preventing Def Leppard's "Pour Some Sugar on Me" from reaching the   \\
Page 2: "Richard Marx (album)" Richard Marx (album) Richard Marx is the debut studio album by singer/songwriter and record producer/arranger, Richard Marx, released in June 1987. Richard became the first male solo artist (and second solo artist overall - the first being Whitney Houston) in recording history to reach the top three of the "Billboard" Hot 100 with four singles from a debut album, including his first number one Pop single "Hold On to the Nights" in mid-1988. The album's first single, "Don't Mean Nothing", hit \#1 on "Billboard"'s Hot Mainstream Rock Tracks chart on July 4, 1987.  \\
Page 3: \textbf{***more search results  omitted here***}
        \\\green{</result>} 
        \\\blue{<think>} Okay, according to the search result, "Hold On to the Nights" is a power ballad by Richard Marx, and it reached number one on the US Billboard Hot 100 chart on July 23, 1988. The result also mentions that it was the fourth and final single from his self-titled debut album. So the answer should be Richard Marx. Let me confirm the time frame. The result says it was in mid-1988, which is within the 80s. So the answer is Richard Marx. But let me check if there's any other artist mentioned in the search results. The third result is about the "Family Guy" episode, but that's not relevant. The second result is about the album, which also states that the first number one was "Hold On to the Nights" in mid-1988. So yes, Richard Marx is the correct answer. \\\blue{</think>}
        \\\purple{<answer>} \\ 
        \boxed{\text{Richard Marx}}
        \\\purple{</answer>}\\
    \bottomrule
    \caption{An example from AT\textsuperscript{2}PO on TriviaQA dataset, with special symbols used in think content, search queries, returned results and final answer highlighted with \blue{blue box}, \red{red box}, \green{green box} and \purple{purple box}, respectively.}
    \label{casestudy6}
    \end{tabular}
\end{table*}

\begin{table*}[!t]
    \centering
    \fontsize{9pt}{11pt}\selectfont
    \begin{tabular}{p{0.98\linewidth}}
    \midrule
        \rowcolor{gray!20}\textbf{Example \#1 from PopQA Dataset} \\
    \midrule
        \textbf{Question:} \\
        What is Eleanor Davis's occupation?\\
        \textbf{Labeled Answer:} \\
        cartoonist/graphic artist/animator/illustrator \\
    \midrule
        \rowcolor{gray!20}
        \textbf{Model Output by AT\textsuperscript{2}PO(Qwen3-8B)}\\
    \midrule
        \blue{<think>} Okay, so I need to find out what Eleanor Davis's occupation is. Let me start by recalling if I know anything about her. Maybe she's an author, artist, or something else? Hmm, not sure. Let me use the Wikipedia search tool to look her up. \blue{</think>}
        \\\red{<search>} What is Eleanor Davis's occupation? \red{</search>}
        \\\green{<result>}\\
Page 1: "Eleanor Layfield Davis" Eleanor Layfield Davis Eleanor Layfield Davis (1911–1985), also called ELDA, was an American painter. She served on the Board of Trustees for Meredith College and both Meredith and Wake Forest University award art scholarships in her memory. Eleanor Layfield Davis was born Eleanor Layfield in 1911, in Richmond, Virginia. She grew up in the Raleigh, North Carolina area, attending Peace College and Sweet Briar College. Davis went on to graduate from Meredith College in 1932. She was married to Egbert Lawrence Davis, Jr. (1911–2006) and had four children. Davis did not start painting until she was 47 years old.         \\
Page 2: "Eleanor Layfield Davis" and in 15 individual exhibitions around the Southeast. Posthumous exhibitions include a retrospective at Wake Forest University's Scales Fine Arts Center Gallery in 1986, and "ELDA – Paintings by Eleanor Layfield Davis" at the Sawtooth Center for Visual Arts in 2012. In early 2010, some of her works were included in a Meredith College alumnae art showcase. Some of Davis artwork was purchased by the North Carolina Art Society for the loan collection of the North Carolina Museum of Art. Eleanor Layfield Davis Eleanor Layfield Davis (1911–1985), also called ELDA, was an American painter.    \\
Page 3: "Eleanor Layfield Davis" primarily signed her works with the name ELDA developed from her initials. In addition to paintings, Davis also drew sketches and sculpted. She spent much of her life helping build up the arts community in Winston-Salem, North Carolina. She studied art at the Arts and Crafts Association of Winston-Salem as well as taking some courses at the University of North Carolina at Greensboro. Davis was a founder and President of Associated Artists of Winston-Salem as well as organizer of Art Gallery Originals. Davis died in 1985. During her lifetime, Davis produced many works that she exhibited in group shows.   
        \\\green{</result>} 
        \\\textbf{***Two turn search steps omitted here***}
        \\\blue{<think>} Oh, there's another result about Eleanor Davis, born in 1983, an American cartoonist and illustrator. So there are two people named Eleanor Davis. The first is Eleanor Layfield Davis, a painter, and the second is Eleanor Davis, a cartoonist. The user might be referring to the cartoonist, given that the search result for "Eleanor Davis" includes that. Let me check which one is more likely. Since the user didn't specify, but the first result is about the painter, but the second is about a cartoonist. Maybe the answer depends on which one is being asked. However, the initial search query was "Eleanor Davis's occupation" and the result includes both. But the second result is more recent, so maybe that's the one. Let me confirm. The second result says Eleanor Davis (born 1983) is an American cartoonist and illustrator. So if the user is referring to that person, the answer is cartoonist and illustrator. But the first result is about a painter. Maybe the user is referring to the cartoonist, as that is more common in modern contexts. However, without more context, it's ambiguous. But since the search result includes both, perhaps the answer is painter. However, the second result is more recent and has more specific info. Let me check the exact name. The result says "Eleanor Davis (born January 16, 1983) is an American cartoonist and illustrator..." So if the user is referring to that person, the answer is cartoonist and illustrator. But the initial search result is about the painter. Maybe the user is referring to the cartoonist. However, the first result is about Eleanor Layfield Davis, which is a different name. Maybe the user made a typo or is referring to the cartoonist. Since the second result is more recent, I think that's the intended person. Let me confirm. Alternatively, maybe the user is referring to the painter. But since the search result includes both, I need to check which one is more likely. Since the user didn't specify, but the first result is about a painter, and the second is about a cartoonist, perhaps the answer is painter. However, the second result is more detailed. Maybe the answer is cartoonist. But without more info, I need to choose based on the search results. Since the second result is from the search, and it's more recent, I think that's the one. So the answer is cartoonist and illustrator. \\\blue{</think>}
        \\\purple{<answer>} \\ 
        \boxed{\text{Richard Marx}}
        \\\purple{</answer>}\\
    \bottomrule
    \caption{An example from AT\textsuperscript{2}PO on PopQA dataset, with special symbols used in think content, search queries, returned results and final answer highlighted with \blue{blue box}, \red{red box}, \green{green box} and \purple{purple box}, respectively.}
    \label{casestudy7}
    \end{tabular}
\end{table*}

\end{document}